\documentclass[10pt,twocolumn,letterpaper]{article}

\usepackage{booktabs}
\usepackage{multirow}
\usepackage[table]{xcolor} %
\usepackage{bm} 
\usepackage{amssymb}
\usepackage{tikz}

\usepackage{wacv}              %

\definecolor{wacvblue}{rgb}{0.21,0.49,0.74}
\usepackage[pagebackref,breaklinks,colorlinks,allcolors=wacvblue]{hyperref}

\def\wacvPaperID{812} %
\def\confName{WACV}
\def\confYear{2026}

\title{RoDyGS: Robust Dynamic Gaussian Splatting for Casual Videos}

\author{
Junmyeong Lee\textsuperscript{*} \quad
Hoseung Choi\textsuperscript{*} \quad
Yoonwoo Jeong\textsuperscript{*} \quad
Minsu Cho\textsuperscript{}
\\[1em]
Pohang University of Science and Technology (POSTECH)
\\
{\tt\small \{junmyeong.lee, hs.choi, jeongyw12382, mscho\}@postech.ac.kr}
\\
}

\begin{document}
\maketitle
\def\thefootnote{*}\footnotetext{Equal contribution.}
\begin{abstract}
4D reconstruction from casually captured monocular videos is challenging due to inherent ambiguity in reconstructing dynamic 3D geometry. 
To address this challenge, we introduce Robust Dynamic Gaussian Splatting (RoDyGS), a method that reconstructs dynamic scene representation from casual monocular videos. RoDyGS explicitly separates static and dynamic scene elements, and applies spatiotemporal regularization to enforce physically plausible geometry and temporally consistent motion.  
Furthermore, we propose a comprehensive benchmark, Kubric-MRig, which provides extensive camera and object motion along with simultaneous multi-view capture, features that are absent in previous benchmarks. Experiments demonstrate that RoDyGS significantly outperforms previous pose-free dynamic novel view synthesis approaches and achieves competitive rendering quality compared to existing pose-free static novel view synthesis approaches. Our proejct page is available at \url{https://rodygs.github.io} 
\vspace{-5mm}
\end{abstract}
    
\section{Introduction}
\label{sec:intro}

\begin{figure}[!tb]
    \centering
    \includegraphics[width=1.0\linewidth]{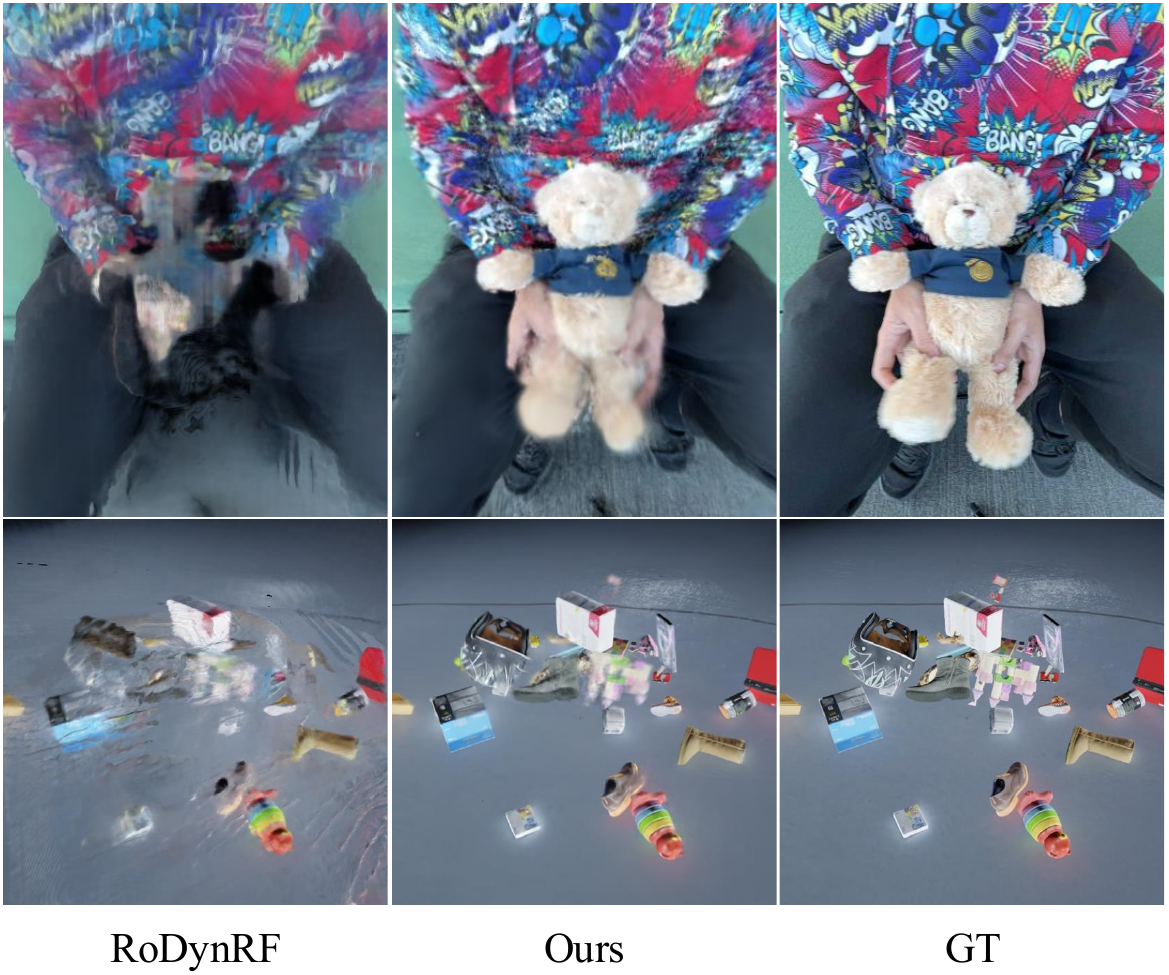}
    \caption{\textbf{Robust Dynamic Gaussian Splatting (RoDyGS).} RoDyGS reconstructs the spatiotemporally consistent geometry of objects and background, rendering high-fidelity dynamic scenes from casually captured monocular videos.}
    \vspace{-5mm}
    \label{fig:main_teaser}
\end{figure}

Everyday environments are filled with complex and dynamic objects, yet we typically capture them using monocular handheld videos. Such videos provide rich visual appearance but lack 3D structure and temporal information, thereby limiting downstream tasks requiring spatial and dynamic scene understanding, such as AR, VR, and robotics.  Accurate 4D reconstruction is therefore essential for recovering complete scene geometry and motion from casual monocular videos. In this paper, we propose a novel end-to-end framework that performs 4D reconstruction directly from monocular videos, without relying on external sensors or manual calibration.

Despite recent advances in 4D reconstruction~\cite{pumarola2020d, park2021nerfies, park2021hypernerf, luiten2023dynamic, wu20244d}, reconstructing accurate scene geometry and motion from videos remains challenging due to two major problems.
First, the ambiguity of camera-object motion complicates the joint estimation of scene geometry and camera trajectory. Second, occlusions result in missing observations, disrupting spatiotemporal consistency in scene elements. To address these issues, recent work has relied on specialized capture setups~\cite{gerats2023dynamic, luiten2023dynamic}, limited scene conditions~\cite{pumarola2020d, park2021hypernerf, wu20244d}, or domain-specific assumptions~\cite{wham:cvpr:2024, guo2023vid2avatar}. While these methods show promising results, they still face limitations when applied to general dynamic scenes casually captured with a monocular camera.

In this work, we propose \textbf{RoDyGS}, a dynamic scene reconstruction method for casually captured monocular videos that jointly optimizes dynamic Gaussian splatting and camera motion. 
A key aspect of RoDyGS is reducing inherent ambiguity in dynamic-scene reconstruction from monocular videos.
We design a part-separation method that explicitly divides static and dynamic scene elements, ensuring stability and enabling part-aware regularization. For dynamic elements, we propose spatiotemporal regularization to preserve geometric consistency and natural motion. Moreover, we jointly optimize camera poses with static regions, ensuring temporally coherent scene representation.

Additionally, existing benchmarks face challenges in evaluating both pose estimation and rendering quality due to the lack of ground truth (GT) camera poses or concurrent multi-view captures. To overcome these limitations, we introduce a challenging benchmark, \textbf{Kubric-MRig}, which includes photorealistic scenes with a variety of simultaneously captured viewpoints and extensive camera and object movement.
The experiments show that RoDyGS outperforms prior pose-free dynamic novel view synthesis methods on Kubric-MRig dataset and achieves competitive performance on iPhone dataset~\cite{gao2022dycheck}. Furthermore, it achieves rendering quality comparable to pose-free static novel view synthesis methods on Tanks and Temples dataset~\cite{knapitsch2017tanks}.  In summary, our contributions are as follows:
\begin{enumerate}
    \item We propose RoDyGS, a robust reconstruction method for casually captured monocular videos that achieves high-fidelity rendering through part separation and spatiotemporal regularization.
    \item We present Kubric-MRig, a challenging dataset designed to evaluate both pose estimation and rendering quality in dynamic scenes, addressing limitations found in existing benchmarks.
    \item Experiments demonstrate that RoDyGS achieves state-of-the-art performance and competitive results across novel view synthesis benchmarks.
\end{enumerate}

\section{Related work}
\label{sec:related_work}
\subsection{Static Novel View Synthesis}

Novel View Synthesis (NVS) is a task of generating novel viewpoints from a set of observations. Pioneering work in NVS leverages point clouds~\cite{kopanas2021point, zhang2022differentiable, xu2022point}, meshes~\cite{riegler2020free, riegler2021stable}, and planes \cite{hoiem2005automatic} for geometrically convincing view synthesis.
Recently, NeRF~\cite{mildenhall2021nerf} has achieved ground-breaking rendering quality by representing volumetric scene functions via MLPs.
To accelerate the training and inference speed, subsequent research focuses on baking trained NeRFs~\cite{hedman2021baking} or directly optimizing explicit representations~\cite{fridovich2022plenoxels, sun2022improved, muller2022instant}.

More recently, 3D Gaussian Splatting (3DGS)~\cite{kerbl20233dgs} has generated novel views by rasterizing anisotropic 3D Gaussians onto image planes. By employing a tile-based alpha-blending CUDA technique, 3DGS achieves real-time rendering with cutting-edge quality on NVS benchmarks.
Building upon 3DGS, several methods have been introduced to enhance fidelity~\cite{kheradmand20243d, ye2024absgs}, enable training with sparse views~\cite{xiong2023sparsegs, zhang2024cor}, and support editing~\cite{chen2024gaussianeditor, dou2024cosseggaussians}. However, these methods rely on pre-calibrated camera poses, limiting their applicability to in-the-wild captures.

To address pose-free scenarios and enhance scalability, iNeRF~\cite{yen2021inerf} estimates camera poses from a pre-trained NeRF by minimizing the discrepancy between the query and rendered views. In subsequent research, \cite{wang2021nerfmm, jeong2021self, lin2021barf, chng2022garf, bian2023nope} optimize both camera poses and NeRF parameters using photometric loss and geometric regularization. Additionally, CF-3DGS~\cite{fu2024cf3dgs} introduces a progressively growing 3DGS for camera pose estimation, while InstantSplat~\cite{fan2024instantsplat} leverages DUSt3R~\cite{wang2024dust3r} to initialize camera poses and point clouds in 3DGS. Yet, these approaches are based on static scenes, which struggle to reconstruct scenes containing typical moving objects found in casual videos.

\subsection{Dynamic Novel View Synthesis}

Following the success of NVS in static scenes, researchers have extended neural fields to capture both the underlying geometry and motion of dynamic scene.
Building on NeRF, \cite{pumarola2020d, park2021nerfies, park2021hypernerf} learn additional time-varying deformation fields.
In parallel, \cite{li2022neural, kplanes_2023, cao2023hexplane} have taken a different approach by focusing on learning multi-dimensional feature fields. These methods encode scene dynamics without relying on explicit motion representations.

With the emergence of 3DGS, \cite{luiten2023dynamic, wu20244d} propose learning the trajectories of individual Gaussians over time.  Subsequent research has introduced more efficient representations, such as factorized motion basis~\cite{kratimenos2024dynmf} and sparse control points~\cite{huang2024sc}. Another line of work~\cite{yang2023gs4d} extends spherical harmonics to a 4D spherindrical harmonics function, integrating both time and view-dependent components.
Despite their reliable performance, these methods assume a limited data distribution for constrained optimization, requiring pre-calibrated poses and functioning under specific conditions, such as camera teleportation or ambient dynamics, thus restricting their applicability to general scenarios in casual videos.

RoDynRF~\cite{liu2023robust} introduces a pose-free optimization method for dynamic radiance fields using a voxelized feature space, enabling reconstruction from casually captured videos. However, it remains limited to forward-facing scenes or small viewpoint changes, restricting its applicability to arbitrary videos.
To address these limitations, we propose RoDyGS, which extends applicability to a wider range of videos through separated optimization and dynamic part-aware regularization techniques.

Moreover, \cite{som2024, wang2024gflow, lee2023fast, shih2024modeling} shares fundamental inspiration to RoDyGS in reconstructing scenes from casual videos. These approaches employ off-the-shelf priors, such as motion masks from the Track Anything Model (TAM)~\cite{yang2023track} or other motion-related cues, to constrain Gaussian optimization. In particular, the methods of \cite{som2024, wang2024gflow, lee2023fast} penalize the 3D motion of individual Gaussians by leveraging external cues generated by long-term point tracking methods~\cite{karaev2023cotracker}. While these priors reduce camera–object ambiguity in simple scenes, they incur extra computational overhead and often fail under occlusions or complex dynamics. In contrast, RoDyGS directly optimizes Gaussian motion within its framework, thereby eliminating these external dependencies.

\section{Preliminary : Dynamic Gaussian Splatting}
\label{sec: method_gaussian_splatting}

\paragraph{Static Representation.} 3DGS represents scene geometry using Gaussian primitives and achieves real-time and high-fidelity rendering through an efficient tile-based rasterization. Specifically, each 3D Gaussian is defined by mean vector \( \bm{\mu}_c \) and 3D covariance matrix \( \bm{\varSigma}_c \). 
According to \cite{ewasplatting}, Gaussians are projected onto the image plane by approximating their 2D means and covariances as follows:
\begin{equation}
\begin{aligned}
    \bm{\mu}^{2D}_c = \bm{\Pi}(KE\bm{\mu}_c), \quad
    \bm{\varSigma}^{2D}_c = \bm{J}\bm{E}\bm{\varSigma}_c\bm{E}^T\bm{J}^T,
\end{aligned}
\end{equation}
where \( \bm{J} \) denotes Jacobian of the affine approximation of the projective transformation, and \( \bm{K} \) and \( \bm{E} \) denote intrinsic and extrinsic matrix of camera, respectively. $\bm{\Pi}$ denotes perspective projection of 3D points into an image plane. Each covariance matrix is decomposed into rotation matrix \( \bm{R}_c \) and scaling matrix \( \bm{S} _c\) such that $\bm{\varSigma}_c = \bm{R}_c\bm{S}_c\bm{S}_c^T\bm{R}_c^T$. In addition, each Gaussian includes opacity $\alpha \in \mathbb{R}$ and spherical harmonics (SH) coefficients $\bm{c} \in \mathbb{R}^{(L+1)^2}$ to represent view-dependent color. 
Thus, the final color of pixel \( \bm{x}_p \) is computed as: 
\begin{equation}
    C_p = \sum_{i=1}^{N} c_i \alpha_i T_i \mathcal{N}\left(\bm{x}_p|\bm{\mu}_c^{2D},\bm{\varSigma}_c^{2D} \right),
\end{equation}
where \(T_i = \prod_{j=1}^{i-1} \left(1-\alpha_j \mathcal{N}\left(\bm{x}_p|\bm{\mu}^{2D}_c,\bm{\varSigma}_c^{2D}\right) \right)\), and \( c_i \) and \( \alpha_i \) represent color and opacity associated with each 3D Gaussian, respectively. 
\paragraph{Dynamic Representation.} 
Our framework adapts Gaussian motion model from DynMF~\cite{kratimenos2024dynmf}, which extends 3DGS to dynamic scenes by representing each Gaussian’s trajectory via learnable \textbf{shared motion bases}.
It predicts $B$-dimensional translation basis $\bm{b}^{\mu}$ and rotation basis $\bm{b}^{q}$ as unit quaternion vectors. Each Gaussian has learnable motion coefficients $\bm{m}$ with dimension $B$, motion of Gaussians is represented by combination of ($\bm{b}^{\mu}$, $\bm{b}^{q}$, $\bm{m}$).
With motion basis function $\bm{\phi}$, we predict motion of Gaussians for timestep $t$ as follows:
\begin{equation}
    (\bm{b}^{\mu}(t), \bm{b}^{q}(t)) = \bm{\phi}(\frac{t}{T}),
\end{equation}
where $\bm{\phi}$ receives the normalized timestep in the range $[0,1]$ by dividing $t$ by maximum timestep $T$. Following \cite{kratimenos2024dynmf}, the motion basis $\bm{b}^{\mu}$ and $\bm{b}^{q}$ are optimized using a set of shallow MLP networks.
Then time-dependent motion parameters $\bm{\mu}(t)$ and $\bm{q}(t)$ are obtained by composing 3D Gaussians' mean $\boldsymbol{\mu}_c$, rotation $\boldsymbol{q}_c$ ($i.e.$, quarternion representation of $\bm{R}_c$), and motion basis function $\boldsymbol{\phi}$, defined as follows: 
\begin{equation}
\begin{aligned}
    \boldsymbol{\mu}(t) = \boldsymbol{\mu}_c + \bm{m} \cdot \boldsymbol{b}^{\mu}(t),  \quad
    \boldsymbol{q}(t) = \frac{\boldsymbol{q}_c + \bm{m} \cdot  \boldsymbol{b}^{q}(t)}{\|\boldsymbol{q}_c + \bm{m} \cdot  \boldsymbol{b}^{q}(t)\|}.
\end{aligned}
\end{equation}

Note that $\bm{q}(t)$ is quaternion representation and can be converted to time-dependent rotation matrix, $\bm{R}(t)$. The rest of rendering process follows the original 3DGS with $\mathbf{R}(t)$ and mean vector $\boldsymbol{\mu}(t)$.

\section{Proposed Method}
\label{sec:method}

In this section, we introduce RoDyGS, an effective method for dynamic Gaussian Splatting from casual monocular videos. 
We first provide an overview of our approach in \Cref{method_overview}. 
In the subsequent sections, we explain our regularization terms to ensure consistent geometry in Section~\ref{geometric_reg} and to capture complex motion in Section~\ref{motion_reg}. 
Finally, we elaborate the details of our optimization in Section~\ref{method_optimization}.

\subsection{Overview of RoDyGS} \label{method_overview} 
\begin{figure*}[!tb]
    \centering
    \includegraphics[width=0.85\textwidth]{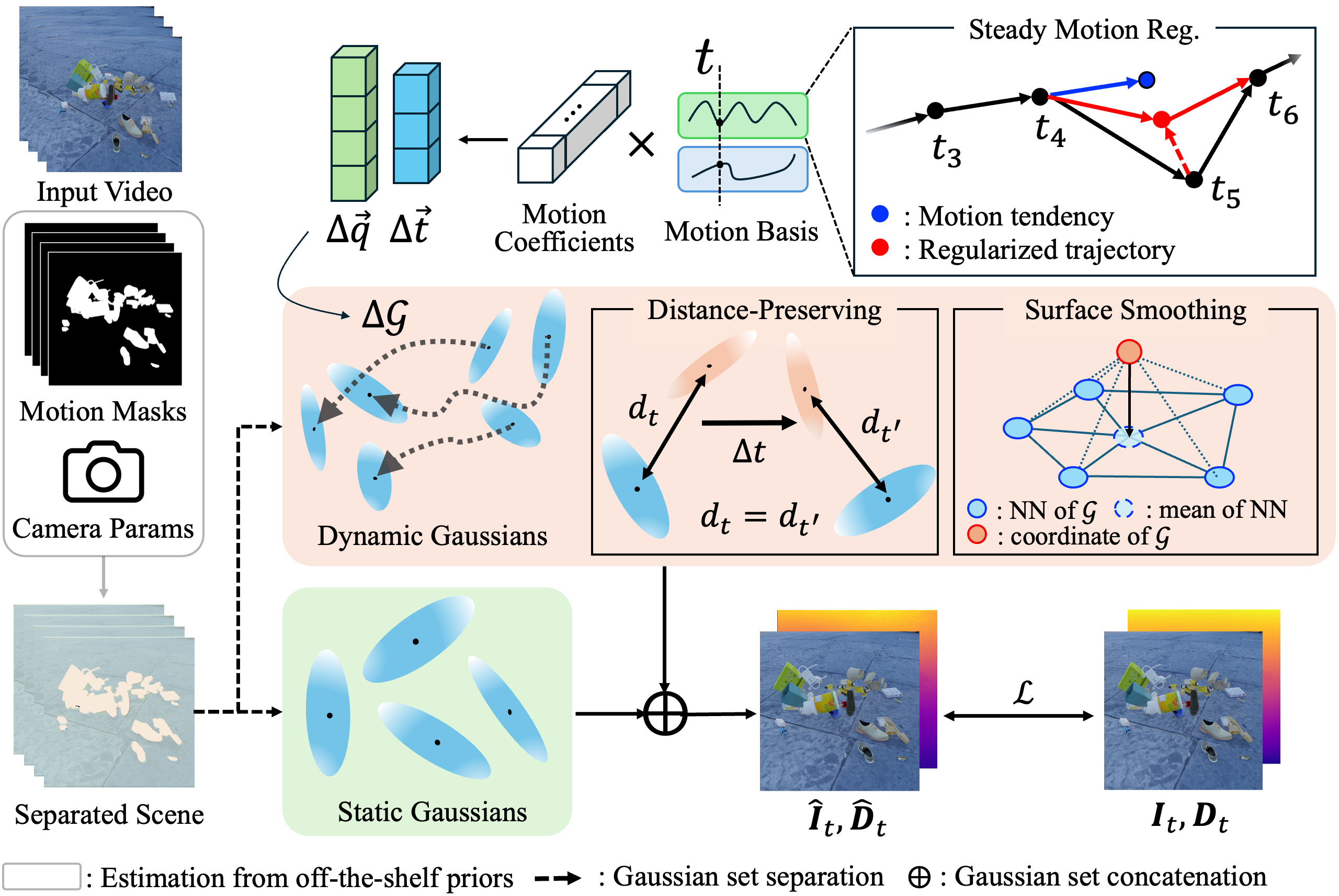}
    \caption{\textbf{Overview of RoDyGS.} 
    Given a casual monocular video, RoDyGS first extracts camera poses using MASt3R~\cite{leroy2024grounding}, while motion masks are derived from TAM~\cite{yang2023track}. It then separates static and dynamic Gaussians, allowing independent optimization for the static and dynamic scene elements.
    The primary optimization objective comprises photometric loss (\( \mathcal{L}_{\text{recon}} \)) and Pearson depth loss (\( \mathcal{L}_{\text{depth}} \)).
    Additionally, for dynamic Gaussians, distance-preserving regularization (\( \mathcal{L}_{\text{dp}} \)) and surface smoothness regularization (\( \mathcal{L}_{\text{surf}} \)) are applied. Steady motion regularization  (\( \mathcal{L}_{\text{sm}} \)) is applied to the motion basis  $\boldsymbol{\phi}$.
    }
    \vspace{-2mm}
    \label{fig:main_overall_pipeline}
    \vspace{-3mm}
\end{figure*}
Given a casual monocular video, RoDyGS reconstructs dynamic scene representation that can synthesize high-quality novel views. It optimizes Gaussian attributes, motion basis $\boldsymbol{\phi}$, and camera parameter $\boldsymbol{E}$ to render a realistic view for a query viewpoint and timestep $t$. We visualize an overview of our method in~\Cref{fig:main_overall_pipeline}.
Before main optimization stage, RoDyGS obtains well-initialized parameters by separating static and dynamic scene elements and estimating initial camera poses.
After initialization, it jointly optimizes camera parameter $\boldsymbol{E}$ and the dynamic scene representation with spatiotemporal regularization terms.
\vspace{-2mm}
\paragraph{Static-Dynamic Part Separation.} We model the scene as combination of static and dynamic scene elements. We partition Gaussians into two distinct sets: set of static Gaussians \( \mathcal{G}_S \), and set of dynamic Gaussians \( \mathcal{G}_D \), where each set represents the corresponding scene elements.
Part separation enhances the flexibility of our method by enabling part-aware objective design.
During optimization, motion-related constraints are selectively applied to $\mathcal{G}_D$, while $\mathcal{G}_S$ are jointly optimized with camera parameter $\boldsymbol{E}$.
This separated optimization disentangles camera motion from dynamic scene elements, significantly reducing camera-object motion ambiguity. 
Additionally, part separation improves rendering efficiency by restricting motion estimation procedure to $\mathcal{G}_D$, thereby reducing computational overhead and accelerating rendering speed.
\vspace{-2mm}
\paragraph{Camera Parameter Refinement.} 
Traditional Structure-from-Motion method~\cite{schoenberger2016mvs} assumes static scene captures, often failing in dynamic environments. 
To address this limitation, we use a data-driven prior model~\cite{leroy2024grounding} to initialize camera parameter $\boldsymbol{E}$ and scene geometry, ensuring robust and accurate camera pose estimation across diverse inputs.
During optimization, we refine camera parameter $\boldsymbol{E}$ with $\mathcal{G}_S$.

\subsection{Geometric Consistency Regularization} \label{geometric_reg}

As discussed in the previous section, 4D reconstruction from a casual monocular video is inherently challenging due to incomplete observations of moving objects, making it difficult to capture their complete geometry and motion. This problem arises from occlusions, as dynamic objects are occluded by other scene elements.
We propose two geometric regularization terms to preserve the geometry of dynamic objects even during occluded intervals in the casual monocular video.
These terms ensure temporal consistency and surface smoothness of the dynamic object geometry.

\vspace{-2mm}
\paragraph{Distance-Preserving Across Time.}
For Gaussian \( g \in \mathcal{G}_D\) and its closest $K$ neighbors \( \text{NN}(g) \), we introduce distance-preserving regularization to enforce the temporal consistency. We define the regularization term $l_{dp}(g,t)$ for Gaussian \( g \) as follows:
\begin{gather}
    d(g_1, g_2, t) = \|\bm{\mu}_{g_1}(t) - \bm{\mu}_{g_2}(t)\|_2^2,  \quad 
    g_1, g_2 \in \mathcal{G}, \\
{\small    l_{\text{dp}}(g, t) = \frac{1}{K |\mathcal{T'}|} \sum_{\substack{g' \in \text{NN(g)} \\ t' \in \mathcal{T'}}}
    \|d(g, g', t) - d(g, g', t')\|_2^2, }
\end{gather}
where \( \bm{\mu}_g(t) \) denotes a mean vector of Gaussian \( g \) at timestep \( t \), and $d(g_1,g_2,t)$ is a distance function between two Gaussians. Using distance $d$, and loss $l_{\text{dp}}(g,t)$, temporal variations in these Gaussian distance across set of sampled timesteps \( \mathcal{T'} \). This constraint preserves the rigidity of the Gaussian set, maintaining geometric consistency even in partially observed regions at certain timesteps.

\vspace{-2mm}
\paragraph{Object Surface Smoothing.} We apply object surface smoothing regularization $l_\text{surf}(g,t)$ to reduce perturbations in the local surface of objects. The motion of individual Gaussians does not inherently preserve surface smoothness, occasionally resulting in locally non-smooth surface over time. To address this, we introduce geometric regularization similar to previous work~\cite{kim2022laplacianfusion}, enforcing each Gaussian to stay near the mean position of its $K$ nearest neighbors $\text{NN}(g)$, defined as follows:
\begin{gather}
    l_{\text{surf}}(g, t) = \|\bm{\mu}_g(t) - \frac{1}{K} \sum_{g' \in \text{NN}(g)}\bm{\mu}_{g'}(t)\|_2^2.
\end{gather}
During training, we combine the two losses into geometric regularization term $\mathcal{L}_{\text{geo}}$. We apply distance-preserving loss \( \mathcal{L}_\text{dp} \) and surface smoothing loss \( \mathcal{L}_\text{surf} \), to randomly sampled dynamic Gaussian set \( \mathcal{G_D'} \subset \mathcal{G}_D \) at timestep \( t \) as follows:
\begin{gather}
    \mathcal{L}_{\text{geo}} = \lambda_{\text{dp}}\mathcal{L}_{\text{dp}} + \lambda_{\text{surf}}\mathcal{L}_{\text{surf}}, 
\end{gather}
where $\mathcal{L}_{*}$ denotes average loss computed over all Gaussians in \( \mathcal{G_D'} \) for corresponding loss function  $l_*(g,t)$. \(\lambda_{\text{dp}}\) and \(\lambda_{\text{surf}}\) are weights for each regularization term.

\subsection{Steady Motion Regularization} \label{motion_reg}

Prior research~\cite{akhter2008nonrigid, kumar2016multi, kratimenos2024dynmf} show that the complex motion of objects can be approximated as low-rank 3D motion. To incorporate this, we first encode a low-rank bias into our motion model by employing learnable motion basis parameterized by bottlenecked MLPs following previous approaches~\cite{ronen2019convergence, heckel2020compressive, kratimenos2024dynmf}. 
Additionally, to explicitly enforce this bias and suppress perturbations in motion basis, we introduce a motion continuity regularization that directly constrains the trajectories derived from motion basis $\boldsymbol{\phi}$.

To effectively learn motion basis that capture real-world dynamics, we impose a \textbf{steady motion assumption}, which assumes consistent motion without drift over short time intervals. Based on the assumption, we regularize motion basis $\boldsymbol{\phi}$ as follows:
\begin{align}
    \textit{l}_{\text{sm}, i}(t) = \|\phi_i(\frac{t}{T}) - \phi_i(\frac{t-1}{T})\|^2_2,
\end{align}
where \( \phi_i \in \bm{\phi} \) is the \(i\)-th element of the motion basis. 
\paragraph{Weighted Motion Regularization.}
While the steady motion assumption effectively models most real-world scenarios, it may not adequately capture complex or fine-grained motion details. To prevent over-smoothing of learned motion, we apply a weighted motion regularization scheme, enabling explicit control over the complexity of each motion basis element $\phi_i \in \boldsymbol{\phi}$. Specifically, we vary the level of regularization applied to each motion basis element by employing a weight generation function. The modified regularization term is defined as follows:
\begin{align}
    \mathcal{L}_{\text{sm}} = \frac{1}{B(T-1)} \sum_{t=2}^T \sum_{i=1}^B w_i l_{\text{sm}, i}(t).
\end{align}

\begin{figure}
    \centering
    \includegraphics[width=0.95\linewidth]{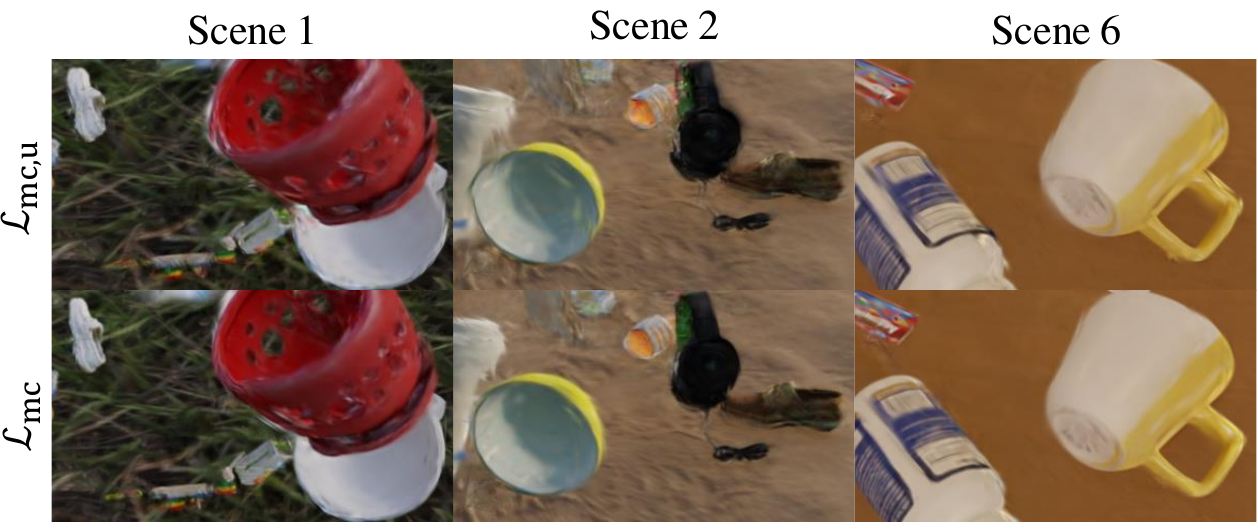}
    \vspace{-2mm}
    \caption{\textbf{Effect of weighted motion regularization on Kubric-MRig dataset.} $\mathcal{L}_{\text{mc, u}}$ (top row) represents motion regularization with uniform weight, $\mathcal{L}_{\text{mc}}$ (bottom row) corresponds to the our weighted motion regularization, which effectively mitigates artifacts near object boundaries.}
    \vspace{-4mm}
    \label{fig:weighted_motion_reg}
\end{figure}
Regularization weight \( w_i \) controls the strength of regularization for each element of motion basis, varying according to a cumulative exponential function:
\begin{equation}
w_i = B \times \left( \frac{1 - \exp(-3i / B)}{\sum_{i}^{B} (1 - \exp(-3i / B))} \right), \quad i = 1, 2, \dots, B.    
\end{equation}

As shown in \Cref{fig:weighted_motion_reg}, the weighted regularization scheme reduces ghosting artifacts and blurring around dynamic object edges compared to uniformly weighted motion regularization.

\subsection{Optimization} \label{method_optimization}
As described in previous section, RoDyGS jointly optimizes Gaussian features, motion basis, and camera parameters with the various regularization terms. 

We use same photometric losses as in the original 3DGS optimization~\cite{kerbl20233dgs}. Specifically, we apply an \( L_1 \) reconstruction loss and an SSIM loss between rendered image $\hat{\textbf{I}}_t$ and target image $\textbf{I}_t$ at timestep $t$: 
\begin{equation}
    \mathcal{L}_{\text{recon}} = \lambda_{l1}(\|\hat{\textbf{I}}_t - \textbf{I}_t\|_1) + \lambda_{\text{SSIM}} (\frac{1 - \operatorname{SSIM}(\hat{\textbf{I}}_t, \textbf{I}_t)}{2}).
\end{equation}

Additionally, we incorporate scene geometry information using monocular depth maps obtained from \cite{depth_anything_v1}. Due to the scale ambiguity of the monocular depth predictions, we apply Pearson depth loss $\mathcal{L}_{\text{depth}}$ used in \cite{xiong2023sparsegs}, instead of distance-based loss. This loss maximizes the linear correlation between rendered depth and estimated affine-invariant depth. 
Thus, our complete objective functions are defined as follows:
\begin{align}
    \mathcal{L}_{\text{static}} &=  \mathcal{L}_{\text{recon}} +\mathcal{L}_{\text{depth}}, \\
    \mathcal{L}_{\text{dynamic}} &= \mathcal{L}_{\text{recon}} +  \mathcal{L}_{\text{depth}} + \mathcal{L}_{\text{geo}} + \mathcal{L}_{\text{sm}}, 
\end{align}
where $\mathcal{L}_{\text{static}}$ and $\mathcal{L}_{\text{dynamic}}$ are total losses of the static and dynamic parts, respectively.

\section{Experiments}
\label{sec:experiments}
\subsection{Dataset: Kubric-MRig} \label{kubric_dataset}
\begin{table}[!t]
    \centering
    \resizebox{\linewidth}{!}{
    \begin{tabular}{lcccc}
        \toprule
        Dataset & Wide Viewpoints & Large Motion & GT Cam & Background \\
        \midrule
        Sintel~\cite{butler2012sintel} & - & \checkmark & \checkmark & \checkmark \\
        T \& T~\cite{knapitsch2017tanks} & \checkmark & - & - & \checkmark \\
        D-NeRF~\cite{pumarola2020d} & \checkmark & \checkmark & \checkmark & - \\
        NVIDIA~\cite{yoon2020novel} & - & \checkmark & \checkmark & \checkmark \\
        HyperNeRF~\cite{park2021hypernerf} & - & \checkmark & - & \checkmark \\
        iPhone~\cite{gao2022dycheck} & - & \checkmark & \checkmark & \checkmark \\
        \rowcolor{gray!20} 
        Kubric-MRig (Ours) & \checkmark & \checkmark & \checkmark & \checkmark \\
        \bottomrule
    \end{tabular}
    }
    \caption{\textbf{Benchmarks for dynamic novel view synthesis.} Previous benchmarks either lack wide viewpoints, large motion, ground truth cameras (GT Cam), or background, whereas Kubric-MRig dataset satisfies all these criteria.}
    \label{tab:main_benchmarks}
    \vspace{-3mm}
\end{table}

\begin{table*}[!ht]
    \centering
    \renewcommand{\arraystretch}{1.1}
    
    \begin{tabular}{lc ccc ccc}
        \toprule
        \multirow{2}{*}{Method} & \multirow{2}{*}{GT Cam} & \multicolumn{3}{c}{Kubric-MRig} & \multicolumn{3}{c}{iPhone~\cite{gao2022dycheck}} \\
        \cmidrule(lr){3-5} \cmidrule(lr){6-8} 
        & & PSNR(↑) & SSIM(↑) & LPIPS(↓) & PSNR(↑) & SSIM(↑) & LPIPS(↓) \\
        \midrule
        D-NeRF~\cite{pumarola2020d} & \checkmark & 20.65 & 0.7160 & 0.4021 & 22.04 & 0.5820 & 0.4952 \\
        RoDynRF~\cite{liu2023robust} & \checkmark & 20.80 & 0.7584 & 0.4836 & 16.78 & 0.5096 & 0.4943 \\
        4DGS1~\cite{wu20244d} & \checkmark & 22.46 & 0.8557 & 0.1760 & 21.25 & 0.5865 & 0.4142 \\
        Deform3D~\cite{yang2024deformable} & \checkmark & 22.66 & 0.8478 & 0.1877 & 22.81 & 0.6950 & 0.3094 \\
        4DGS2~\cite{yang2023gs4d} & \checkmark & 23.02 & 0.8460 & 0.1898 & 25.52 & 0.7843 & 0.2535 \\
        \midrule
        RoDynRF~\cite{liu2023robust} & - & 18.36 & 0.6121 & 0.5984 & 14.91 & 0.4050 & 0.5531 \\ 
        SoM~\cite{som2024} & - & 16.83 & 0.6283 & 0.6682 & \textbf{17.55} & 0.4569 & 0.4679 \\ 
        \rowcolor{gray!20} 
        \textbf{RoDyGS (Ours)} & - & \textbf{19.44} & \textbf{0.7169} & \textbf{0.3401} & 17.38 & \textbf{0.4656} & \textbf{0.4361} \\
        \bottomrule
    \end{tabular}
    
    \caption{\textbf{Dynamic novel view synthesis quality on Kubric-MRig and iPhone datasets.} 
    GT Cam denotes the availability of ground truth camera poses during training. 
    Our method achieves superior performance over RoDynRF for novel view rendering quality.}
    \label{tab:main_exp_kubric_dynamic}
    \vspace{-3mm}
\end{table*}

We revisit previous Dynamic Novel View Synthesis benchmarks---Sintel~\cite{butler2012sintel}, Tanks and Temples~\cite{knapitsch2017tanks}, D-NeRF~\cite{pumarola2020d}, NVIDIA Dynamic~\cite{yoon2020novel}, HyperNeRF~\cite{park2021hypernerf}, and iPhone~\cite{gao2022dycheck}---to assess their suitability with our problem. As summarized in Table~\ref{tab:main_benchmarks}, the existing benchmarks have several limitations in evaluating camera pose estimation and view synthesis performance in causal monocular video scenarios: they offer restricted viewpoints such as forward-facing scenes~\cite{butler2012sintel, gao2022dycheck, yoon2020novel, park2021hypernerf}, feature no motion~\cite{knapitsch2017tanks}, lack ground truth camera poses~\cite{knapitsch2017tanks, yoon2020novel}, and miss background~\cite{pumarola2020d}. To address these issues, we introduce Kubric-MRig, which includes large movements of both the camera and objects, ground truth camera poses, and background.

To build Kubric-MRig dataset, we use Kubric~\cite{greff2021kubric}, a Blender-based synthetic scene generator. Each scene consists of two 100-frame monocular videos: one for training and one for testing. The training video is captured by moving the camera around the objects. Following the evaluation protocol of \cite{yoon2020novel}, the camera that captures the testing video is fixed at the initial position of the training video and captures 100 viewpoints throughout its duration. More details are provided in the supplementary.

\subsection{Implementation Details}
We incorporate pre-trained priors for well-initializated parameters and depth maps. Specifically, we use MASt3R~\cite{leroy2024grounding} to initialize camera poses and point clouds for initial Gaussians, and use DepthAnything~\cite{depth_anything_v1} to estimate monocular depth maps.
Our method does not require scene-specific hyperparameter tuning, allowing for seamless application to casual monocular videos without the need for additional tuning procedures.

Additonally, we found that the dynamic mask generation procedure of RoDynRF, which utilizes RAFT~\cite{teed2020raft} flow and Mask-RCNN~\cite{he2017mask}, fails to produce reliable motion masks for Kubric-MRig dataset.
Therefore, we use TAM~\cite{yang2023track} for more stable motion mask generation. We report an in-depth comparison between the dynamic masks of RoDynRF and TAM in the supplementary material.

\subsection{Dynamic View Synthesis}

\begin{figure*}[!t]
    \centering
    \includegraphics[width=1.0\linewidth]{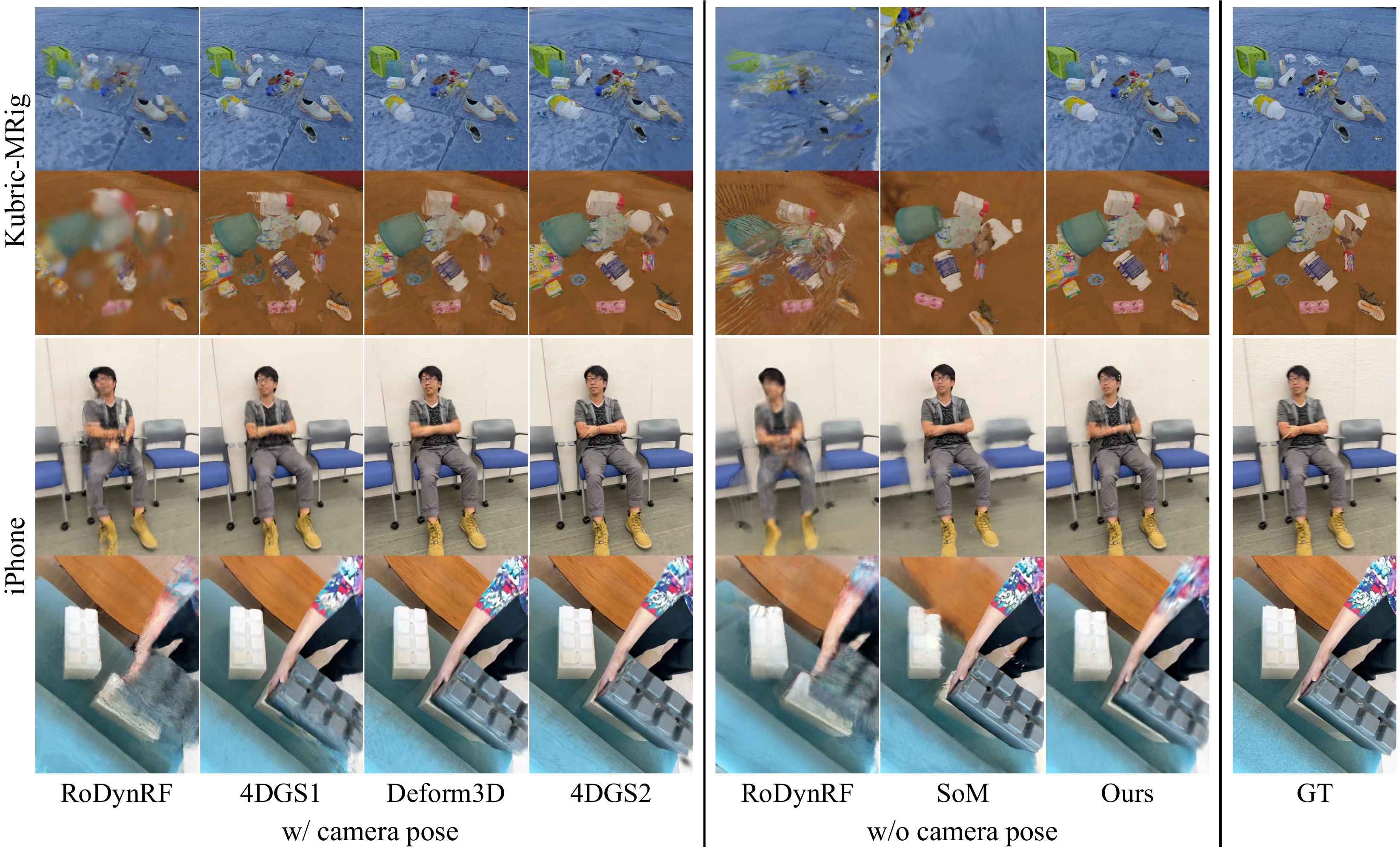}
    \caption{\textbf{Qualitative results on Kubric-MRig and iPhone dataset.} We render dynamic novel views from casual monocular videos. Our method accurately reconstructs scene geometry, produces sharp renderings, and aligns object positions well compared to other baselines.
    }
    \label{fig:main_qual_mrig_dynamic}
    \vspace{-2mm}
\end{figure*}

We compare RoDyGS with previous dynamic novel view synthesis approaches that are trained with and without ground truth camera poses~\cite{pumarola2020d, liu2023robust, yang2023gs4d, wu20244d, yang2024deformable, som2024}. 
We assess the dynamic novel view synthesis quality with PSNR, SSIM, and LPIPS. We conduct experiments on Kubric-MRig and iPhone dataset~\cite{gao2022dycheck}.
For Kubric-MRig dataset, we follow the evaluation protocol of \cite{yoon2020novel}, measuring dynamic novel view rendering performance at the initial training camera pose.
For iPhone dataset, the standard evaluation protocol assumes that ground truth camera poses are provided for training~\cite{gao2022dycheck}, whereas our method does not require known camera poses. To ensure a fair comparison, we reorganize the training and test splits from a single camera trajectory, enabling an accurate evaluation.
Please refer to the supplementary material for more details on the evaluation protocols.

Our quantitative results in \Cref{tab:main_exp_kubric_dynamic} show that RoDyGS outperforms camera pose–free methods~\cite{liu2023robust, som2024} on Kubric-MRig dataset and achieves competitive results on iPhone dataset.
Qualitatively, as shown in \Cref{fig:main_qual_mrig_dynamic}, RoDyGS produces clearer renderings than camera pose-free baselines and attains visual fidelity on par with methods optimized using ground-truth camera parameters. Specifically, our approach more effectively preserves geometric accuracy and temporal consistency, yielding sharp object boundaries without the shape distortions or ghosting artifacts exhibited by other baselines. In Kubric-MRig dataset (rows 1–2 of \Cref{fig:main_qual_mrig_dynamic}), RoDyGS maintains rigid object geometry even under large camera and object motion, whereas other baselines show inconsistent reconstruction.

We also observe that methods relying on long-term point-tracking priors often struggle under large camera motion and in complex dynamic scenes due to inaccurate point tracks, as illustrated by the results of SoM~\cite{som2024} in the top row of \Cref{fig:main_qual_mrig_dynamic}. Additional comparisons on Kubrig-MRig dataset and detailed experimental results are provided in the supplementary material.

\begin{table}[]
    \centering
    \renewcommand{\arraystretch}{1.1}
    
    \vspace{-2mm}
    \resizebox{1.0\linewidth}{!}{
    \begin{tabular}{lccccccc}
        \toprule
         Method & PSNR($\uparrow$) & SSIM($\uparrow$) & LPIPS($\downarrow$) & ATE($\downarrow$) & RPE-R($\downarrow$) & RPE-t($\downarrow$) \\
        \midrule
        NeRFmm~\cite{wang2021nerfmm} & 22.50 & 0.59 & 0.54 & 0.123 & 0.477 & 1.735 \\
        SC-NeRF~\cite{jeong2021self} & 23.76 & 0.65 & 0.48 & 0.129 &  0.489 & 1.890  \\
        BARF~\cite{lin2021barf} & 23.42 & 0.61 & 0.54 & 0.078 & 0.441 & 1.046 \\
        Nope-NeRF~\cite{bian2023nope} & 26.34 & 0.74 & 0.39 & 0.006 & \textbf{0.038} & 0.080 \\
        CF-3DGS~\cite{fu2024cf3dgs} & 31.28 & \textbf{0.93} & 0.09 & \textbf{0.004} & 0.069 & \textbf{0.041} \\ 
        \rowcolor{gray!20} 
        \textbf{RoDyGS (Ours)} & \textbf{31.51} & \textbf{0.93} & \textbf{0.08} & 0.007 & 0.096 & 0.052 \\
        \bottomrule
    \end{tabular}
    }
    \vspace{-2mm}
    \caption{\textbf{Comparison of pose-free static novel view synthesis methods on Tanks and Temples dataset.} We evaluate novel view rendering and pose estimation performance on Tanks and Temples dataset with pose-free static novel view synthesis approaches. RoDyGS achieves competitive performance with CF-3DGS.}
    \vspace{-5mm}
    \label{tab:main_exp_tanks}
\end{table}
\subsection{Static Novel View Synthesis}

To comprehensively evaluate performance on static scenes, we compare RoDyGS with existing pose-free static novel view synthesis methods~\cite{wang2021nerfmm, jeong2021self, lin2021barf, bian2023nope, fu2024cf3dgs}. The evaluation is performed on Tanks and Temples dataset~\cite{knapitsch2017tanks}, following the protocol of CF-3DGS~\cite{fu2024cf3dgs}. As reported in \Cref{tab:main_exp_tanks}, RoDyGS achieves rendering quality comparable to CF-3DGS and significantly outperforms the other baselines. In camera pose estimation, RoDyGS attains accuracy on par with CF-3DGS and Nope-NeRF~\cite{bian2023nope}, while substantially reducing optimization time and memory requirements.

\subsection{Component Analysis of RoDyGS}
\paragraph{Effect of the Off-the-Shelf Priors.}
For a fair comparison with baseline, we incorporate the same off-the-shelf priors used to RoDynRF optimization. Specifically, we apply two RoDynRF variants, using TAM~\cite{yang2023track} object masks and camera parameters from MASt3R~\cite{leroy2024grounding} as optimization priors.
Quantitative results on iPhone dataset~\cite{gao2022dycheck} are presented in \Cref{tab:rodynrf_variation_exp}.
Note that we exclude the “block” scene from the evaluation due to failure in optimizing the RoDynRF variant with MASt3R.

Compared to original RoDynRF, using TAM masks instead of RoDynRF masks results in degraded performance. We attribute this to the larger TAM masks, which likely introduce blurriness in dynamic regions during rendering. Initializing RoDynRF with MASt3R~\cite{leroy2024grounding} camera parameters improves performance across all metrics but still underperforms our method. 
Overall, \Cref{tab:rodynrf_variation_exp} shows that RoDyGS consistently outperforms all RoDynRF variants, indicating that RoDyGS's performance arises not merely from better priors, but from our robust reconstruction which optimizes both static and dynamic scene elements. Additional experimental results on camera pose priors provided in supplementray material.

\paragraph{Ablation Study.}

\begin{figure}[!t]
    \centering
    \includegraphics[width=\linewidth]{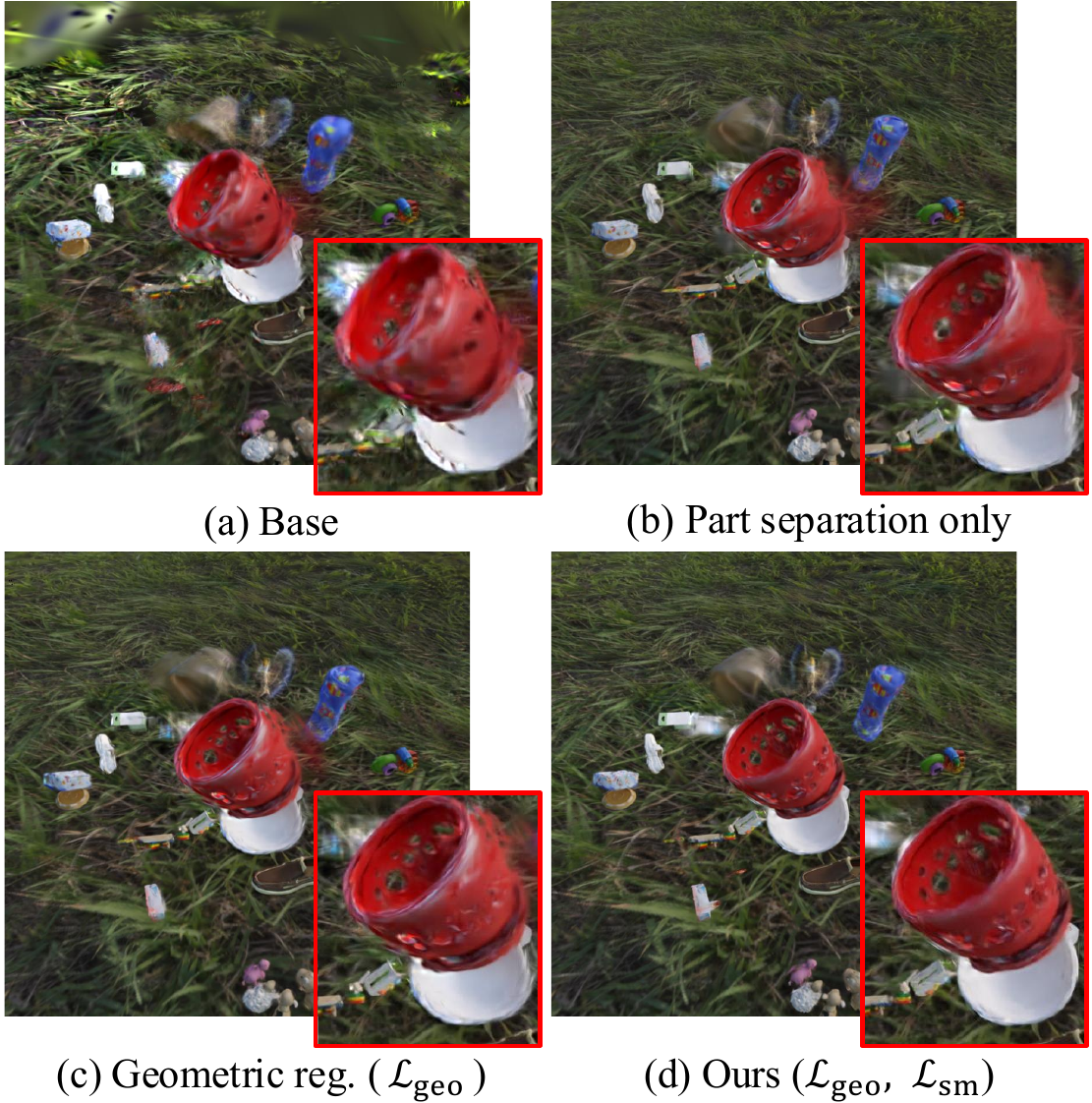}
    \caption{\textbf{Rendering results of the ablation study.} Our regularization effectively enhances the perceptual quality of the rendering results, leading to sharper and more realistic renderings. 
    }
    \label{fig:reg_ablation}
    \vspace{-3mm}
\end{figure}
We conduct ablation study on the part separation and regularization terms using Kubric-MRig dataset. Our quantitative experimental results are reported in \Cref{tab:main_ablation_reg}. We compare our model under the following settings: (a) without part separation and any regularization, (b) only part separation, (c) part separation with $\mathcal{L}_{\text{geo}}$, and (d) part seperation anthe full RoDyGS model.

The results in \Cref{tab:main_ablation_reg} demonstrate the effectiveness of our design. As discussed earlier, part separation significantly improves optimization stability, leading to substantial gains in quantitative metrics. Moreover, our regularization terms consistently improve view rendering performance, with $\mathcal{L}_\text{sm}$ largely improving rendering quality, as reflected in quantitative evaluations.

As shown in \Cref{fig:reg_ablation}, our approach significantly improves perceptual rendering quality. Without part separation, we observe severe inconsistency across timesteps, with objects losing shape and colors appearing distorted. Introducing part separation notably improves background consistency and results in more natural object motion. Adding geometric regularization further sharpens object shapes, but it may occasionally introduce minor artifacts near object boundaries, as illustrated by the edges of the red basket in \Cref{fig:reg_ablation}-(c). Finally, our steady motion regularization effectively maintains the structural consistency of objects, yielding renderings with clearly visible object shapes and sharper boundaries.

\section{Limitations and Conclusion}
\label{sec:conclusion}
\paragraph{Limitations.} Although our approach achieves promising results for casually captured monocular videos, several limitations remain. Similar to other 4D reconstruction methods, RoDyGS only reconstruct the observed parts of scene elements in input videos. If certain parts are completely occluded throughout the entire sequence or experience long-term occlusions, RoDyGS fails to recover them. We hypothesize that this issue would be addressed by leveraging generative data-driven priors, such as diffusion models, to infer unobserved regions.
Additionally, RoDyGS encodes a weak rigidity constraint in the regularization terms, which assumes that dynamic scene elements do not exhibit highly complex and non-rigid deformations, such as those found in humans or loose clothing. 
To generalize our approach to everyday videos, we need to incorporate complex object deformations through parametric models (e.g., SMPL for human dynamics) or other data-driven priors.

\vspace{-4mm}
\paragraph{Conclusion.}
In this work, we propose RoDyGS, a novel and robust 4D reconstruction method for casually captured monocular videos. We introduce part separation and regularization terms to capture complex scene motion as well as the underlying geometry.
For rigorous evaluation, we present Kubric-MRig, a comprehensive benchmark designed to challenge existing dynamic view synthesis methods. This benchmark features significant object motion and wide camera baselines, as well as simultaneous multi-view sequences—key characteristics for evaluating the limits of dynamic scene modeling.
Experimental results show that RoDyGS significantly outperforms prior pose-free dynamic novel view synthesis methods and achieves results comparable to previous pose-free static novel view synthesis methods.
Moreover, our ablation study show that the part separation and regularization terms effectively enforce the learning of realistic motion and temporally consistent geometry.

\begin{table}[!t]
    \centering
    \resizebox{0.9\linewidth}{!}{
    \begin{tabular}{lccc}
        \toprule
         Method & PSNR(↑) & SSIM(↑) & LPIPS(↓) \\
        \midrule
         RoDynRF~\cite{liu2023robust} & 14.91 & 0.4050 & 0.5531 \\
         \ \ \ \ \ + TAM~\cite{yang2023track} & 13.82 & 0.3535 & 0.5336   \\
         \ \ \ \ \ + MASt3R \cite{leroy2024grounding} & 15.40 & 0.3649 & 0.5798 \\
        \midrule
        \rowcolor{gray!20} 
        
        \textbf{RoDyGS (Ours)} & \textbf{17.33} & \textbf{0.4537} & \textbf{0.4296} \\
        \bottomrule
        \end{tabular}
        }
    \vspace{-2mm}
    \caption{\textbf{Evaluation result of RoDynRF with off-the-shelf priors on iPhone dataset.} (Exclude "block" scene.)}
    \label{tab:rodynrf_variation_exp}
    \vspace{-2mm}
\end{table}

\begin{table}[!t]
    \centering
    \resizebox{\linewidth}{!}{
    \begin{tabular}{rccc ccc}
        \toprule
        & part sep. & $\mathcal{L}_{\text{geo}}$ & $\mathcal{L}_{\text{sm}}$ & PSNR($\uparrow$) & SSIM($\uparrow$) & LPIPS($\downarrow$) \\
         \midrule
        (a) & - & - & - & 16.41 & 0.5378 & 0.5172 \\
        
        (b) & \checkmark & - & - & 19.05 & 0.7063 & 0.3525 \\
        (c) & \checkmark & \checkmark &-  & 19.08 & 0.7122 & 0.3481 \\
        \rowcolor{gray!20} 
        
        (d) & \checkmark & \checkmark& \checkmark &\textbf{19.44} & \textbf{0.7169} & \textbf{0.3401} \\
        \bottomrule
        \end{tabular}
        
    }
    \vspace{-2mm}
    \caption{\textbf{Ablation study on Kubric-MRig dataset.} Each row corresponds to images in \Cref{fig:reg_ablation}.}
    \label{tab:main_ablation_reg}
    \vspace{-4mm}
\end{table}

{
    \small
    \bibliographystyle{ieeenat_fullname}
    \bibliography{main}
}

\clearpage
\maketitlesupplementary
\setcounter{page}{1}
\setcounter{section}{0}

\renewcommand{\thesection}{\Alph{section}}
\renewcommand{\thesubsection}{\Alph{section}.\arabic{subsection}}

\renewcommand{\theHsection}{supple.\Alph{section}}
\renewcommand{\theHsubsection}{supple.\Alph{section}.\arabic{subsection}}

\section{Evaluation Data and Protocol}

\subsection{Generation of Kubric-MRig Dataset}
We provide a detailed explanation of the generation process for our Kubric-MRig dataset.
Note that our generation process is implemented based on the Movi script, as introduced in Kubric~\cite{greff2021kubric}.
A primary concern in our dataset design is to mitigate potential ambiguity between camera and object motion, particularly when dynamic objects are excessively more prominent than static ones.

To construct each scene, we begin by selecting 15 to 35 objects from the Google Scanned Objects~\cite{downs2022google} dataset. 
We then place 10 to 20 of these objects on the ground plane, while positioning the remaining objects suspended in the air. 
To achieve natural-looking motion, we apply physics simulations to the dynamic objects, ensuring their movements closely resemble real-world dynamics. 
We point out that the objects on the ground plane exhibit dynamic motion due to collisions with each other, resulting in complex and realistic dynamics for each scene.

\subsection{Evaluation Protocol}
\paragraph{Tanks and Temples Dataset.}
For evaluation on the Tanks and Temples dataset~\cite{knapitsch2017tanks}, we follow the same protocol as CF-3DGS~\cite{fu2024cf3dgs}.
In detail, we use 12.5\% of the frames for evaluation, while the remaining frames are used for training.
All remaining images are used for training.

Since pose-free static novel view synthesis approaches are trained without camera pose information, an additional alignment step is required for evaluation on novel viewpoints.
During evaluation, we first align all test camera poses to the training camera poses before computing metrics.
Specifically, we initialize each test camera pose by finding its nearest neighbor among the calibrated training camera poses, then refine it by minimizing the photometric error between the rendered and ground truth test images.
To enable fair comparison between RoDyGS and static pose-free neural fields, we disable all motion detection by setting all motion masks to zero when evaluating on static scenes.
\vspace{-3mm}
\paragraph{iPhone Dataset.}
The iPhone dataset~\cite{gao2022dycheck} was originally designed for evaluating pose-aware novel view synthesis methods.
However, when evaluating pose-free dynamic novel view synthesis approaches, we found that the large distances between test and training cameras made it difficult to accurately align test camera poses.
To enable proper evaluation of pose-free neural fields, we modify the dataset split to follow CF-3DGS~\cite{fu2024cf3dgs}'s protocol, which ensures test cameras remain close enough to training cameras for reliable pose alignment.

\begin{figure*}[!htb]
    \centering
    \includegraphics[width=1\textwidth]{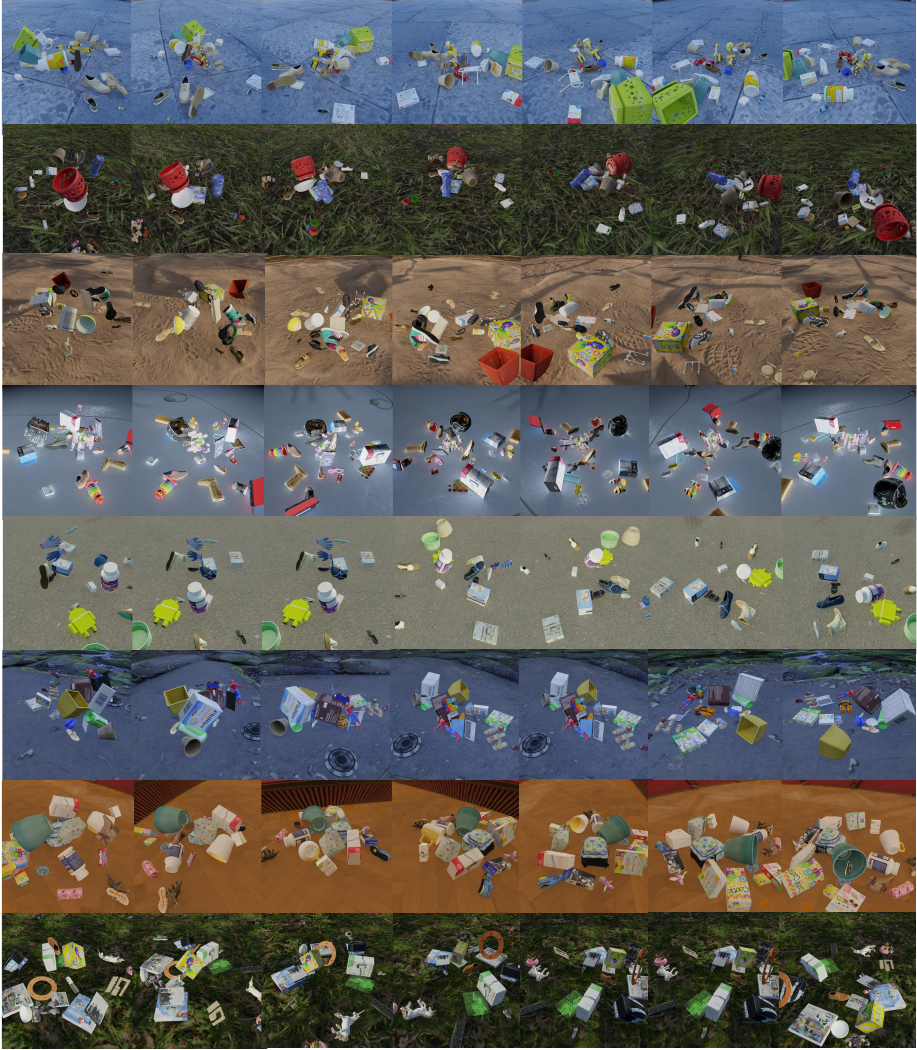}
    \caption{\textbf{Samples from Kubric-MRig dataset.} Kubric-MRig is a dataset generated using Blender that contains 8 scenes. Each scene features multiple objects, some static and some in motion.}
    \label{fig:appendix_dataset}
    \clearpage
\end{figure*}

We evaluate on seven scenes from iPhone dataset: \textit{apple, block, paper-windmill, space-out, spin, teddy,} and \textit{wheel}.
For each scene, we use the first camera 0 (the training camera in the original dataset) to create our train/test split.
Specifically, we use 12.5\% of the frames for evaluation, while the remaining frames are used for training.
This sampling strategy ensures that the test frames are well-distributed throughout the sequence while maintaining sufficient proximity to training frames.
Following the evaluation protocol used in CF-3DGS, this proximity allows us to robustly align the test camera poses during evaluation.
\vspace{-3mm}
\paragraph{Kubric-MRig Dataset.} 
For training, we deploy 100 cameras that follow circular trajectories around the objects. 
Then, we follow the evaluation protocol proposed in \cite{yoon2020novel} in our Kubric-MRig dataset to construct the evaluation setup.
In detail, we fix the camera viewpoint to the first camera used during training and capture the scene across the 100 timesteps. 
All cameras are positioned equidistant from the world center, with distances randomly sampled between 15 and 20 units. 
To ensure comprehensive viewpoint coverage of the scenes for evaluation, we fix the elevation angle, which is randomly sampled from \(30^\circ\) to \(60^\circ\) during data capture. 
We provide a visualization of the generated dataset in Figure~\ref{fig:appendix_dataset}.

\section{Implementation Details}
\subsection{Leveraging Off-the-shelf Priors}

Our optimization process begins with two initialization steps.
First, we use MASt3R~\citep{leroy2024grounding} to obtain initial scene geometry, using the ‘swin’ option with a window size of 10 and running it for 200 iterations.
Second, we employ the Track Anything Model (TAM)~\citep{yang2023track} to identify moving objects in the scene.
By leveraging motion masks obtained from TAM, we separately initialize static and dynamic Gaussians.
To initialize the point clouds for Gaussian Splatting, we uniformly sample 120,000 points from the unprojected point clouds across all frames.
For additional geometric supervision, we incorporate monocular depth estimates from DepthAnything~\cite{depth_anything_v1}.

\subsection{Learnable Motion Basis}

We employ learnable motion basis representation implemented using a set of shallow MLP networks, as proposed in DynMF~\cite{kratimenos2024dynmf}. 
The architecture consists of \( B \) learnable motion bases, structured as a shared time encoder and \( B \) motion prediction heads. 
In our experiments, we use the number of basis, $B$, of 16. 
The encoder is a 3-layer MLP, with input timestep expanded using sinusoidal embeddings at 26 frequency variations.
It has a hidden layer with a width of 128 and uses the GELU activation function.
Each motion prediction head is a 2-layer MLP with a width of 64, without activations in the final layer.

Additionally, we adopt Gaussian motion coefficient regularization from DynMF, incorporating the sparsity regularization losses \( \mathcal{L}_{\text{m}} \) and \( \mathcal{L}_{\text{ms}} \). 
Defining the motion coefficient vector \( \bm{m}_i \) for the \( i \)-th Gaussian, these sparsity losses promote sparsity in the motion coefficient vector, thereby reducing overfitting and mitigating the impact of noisy motion in the training viewpoints.
They are formally defined as:
\begin{gather}
\mathcal{L}_{\text{m}} = \frac{1}{N B} \sum_{i=1}^N \sum_{j=1}^B \|m_{ij}\|_1, \\ 
\mathcal{L}_{\text{ms}} = \frac{1}{N} \sum_{i=1}^N \left( \frac{1}{B} \sum_{j=1}^B \frac{\|m_{ij}\|_1}{\max\limits_{1 \leq k \leq B} \|m_{ik}\|_1} \right).
\end{gather}
Note that these motion sparsity losses are also applied to our optimization process.

For ablation study, we reproduce DynMF by adding motion coefficient regularization instead of our proposed regularization terms. 
This loss is defined for the motion coefficients of  \( i \)-th and \( j \)-th Gaussians, \( \bm{m}_i \) and \( \bm{m}_j \), respectively:
\begin{small}
\begin{equation}
        \mathcal{L}_{\text{coeff}} = \frac{1}{Nk} \sum_{i=1}^N \sum_{j \in \text{NN}(\mathcal{G}_i)} \exp\left( -\lambda_w \| \bm{\mu}_i - \bm{\mu}_j \|^2_2 \right) \| \bm{m}_i - \bm{m}_j \|^2_2.
\end{equation}
\end{small}

$\mathcal{L}_{\text{coeff}}$ is applied to the \( k \) nearest neighbors of the \( i \)-th Gaussian, \( \mathcal{G}_i \). Therefore, the total motion loss, originally proposed in DynMF is expressed as:
\begin{align}
\mathcal{L}_{\text{motion}} = \lambda_{\text{coeff}} \mathcal{L}_{\text{coeff}} + \lambda_{\text{m}} \mathcal{L}_{\text{m}} + \lambda_{\text{ms}} \mathcal{L}_{\text{ms}},   
\end{align}
where \( \lambda_{\text{coeff}} \), \( \lambda_{\text{m}} \), and \( \lambda_{\text{ms}} \) are hyperparameters that control the contributions of each loss term.

\subsection{Optimization Details}

\paragraph{Pearson Depth Loss.} We incorporate scene geometry information using monocular depth maps obtained from DepthAnything~\cite{depth_anything_v1}. However, due to the scale ambiguity of the predicted monocular depths, we cannot directly compare these predictions with the rendered scene depth from the Gaussians. To address this issue, we apply the Pearson depth loss \( \mathcal{L}_{\text{depth}} \)~\cite{xiong2023sparsegs}, which maximize the linear correlation between the rendered depth and the estimated depth. Specifically, \( \mathcal{L}_{\text{depth}} \) is designed to maximize the linear correlation between the rendered depth map \( \hat{\boldsymbol{D}}_t \) and the estimated depth \( \boldsymbol{D}_t \), as follows:
\begin{align}
    \mathcal{L}_{\text{depth}} &= \frac{1}{N} \sum_{t=1}^{N_F} \left( 1 - \mathcal{E}(\hat{\boldsymbol{D}}_t, \boldsymbol{D}_t) \right), \\
    \mathcal{E}(\hat{\boldsymbol{D}}_t, \boldsymbol{D}_t) &= \frac{\mathbb{E}[\hat{\boldsymbol{D}}_t \boldsymbol{D}_t] - \mathbb{E}{[\hat{\boldsymbol{D}}_t}] \mathbb{E}[\boldsymbol{D}_t]}{\sigma[\hat{\boldsymbol{D}}_t] \cdot \sigma[\boldsymbol{D}_t]},
\end{align}
where $\sigma$ is the standard deviation. We compute global depth loss $\mathcal{L}_{\text{depth, g}}$ to match global depth maps and local depth loss $\mathcal{L}_{\text{depth, l}}$ to compare local statistics, which remove local noise of depth, as follows:
\begin{equation}
    \mathcal{L}_{\text{depth}} = \lambda_{\text{depth, g}}\mathcal{L}_{\text{depth, g}} + \lambda_{\text{depth, l}}\mathcal{L}_{\text{depth, l}}.
\end{equation}

\paragraph{Parameter Settings.}
We use the official implementation of 3D Gaussian Splatting (3DGS), with modifications to enable gradient computation over camera poses. 
To avoid overfitting and stabilize the optimization process, we maintain the spherical harmonic (SH) coefficient degree of the Gaussians at 0. 
Afterward, we further optimize our model for 5000 steps, gradually increasing the SH coefficient degree every 1000 iterations, up to a maximum degree of 3.
The weights assigned to each loss term during the optimization process are detailed in Table~\ref{tab:loss_weights}.
Note that $\lambda_{\text{depth,g}}$ and $\lambda_{\text{depth,l}}$ are applied to Pearson depth loss calculated from the rendered global depth map and the local depth maps which are obtained by dividing the global depth map into local patches, respectively.

\begin{table}[h]
\centering
\begin{tabular}{ll}
\toprule
\textbf{Loss term}       & \textbf{Weight} \\
\midrule
$\lambda_{\text{SSIM}}$  & 0.2             \\
$\lambda_{l1}$           & 0.8             \\
$\lambda_{\text{depth, g}}$ & 0.05          \\
$\lambda_{\text{depth, l}}$ & 0.15          \\
$\lambda_{\text{dp}}$           & 0.5             \\
$\lambda_{\text{surf}}$            & 0.5             \\
$\lambda_{\text{sm}}$           & 0.1             \\
$\lambda_{\text{m}}$            & 0.05           \\
$\lambda_{\text{ms}}$           & 0.002            \\
\bottomrule
\end{tabular}
\caption{Weights are assigned to each loss term during the optimization process.}
\label{tab:loss_weights}
\end{table}

We use a linear warm-up strategy for learning rates during the first 10\% of the total training steps, followed by cosine annealing for the rest of training steps. 
The peak learning rates are set to $1.0 \times 10^{-5}$ for camera rotation and $1.0 \times 10^{-6}$ for camera translation.

\section{Discussion Regarding Motion Masks}

\subsection{Obtaining Motion Mask with Epipolar Errors} \label{epipolar_motion_mask}

\begin{figure}[t]
    \centering

    \begin{subfigure}[t]{0.32\linewidth} 
        \centering
        \includegraphics[width=\textwidth]{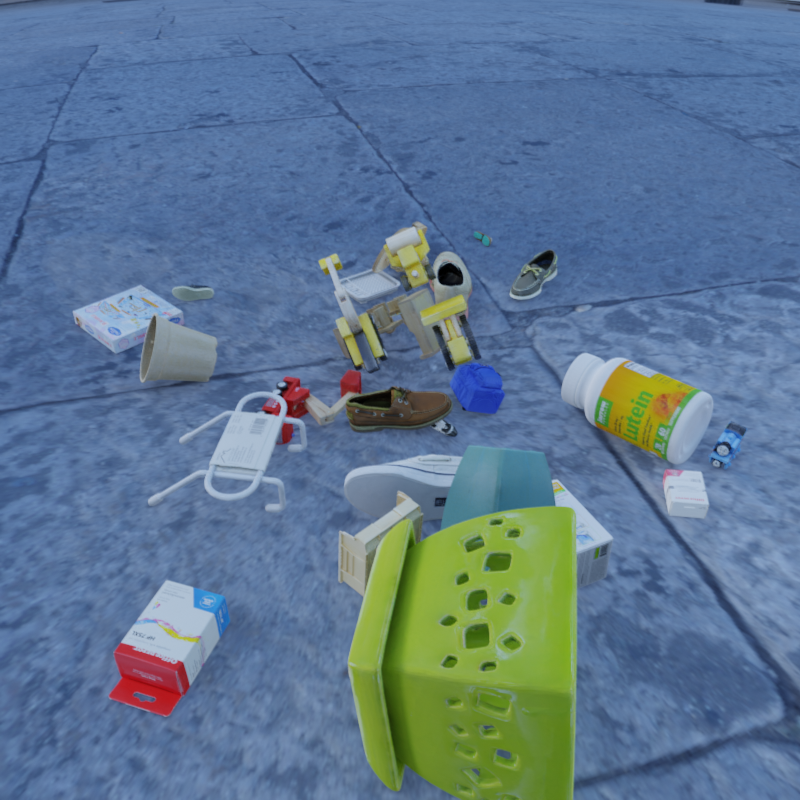} \\
        \includegraphics[width=\textwidth]{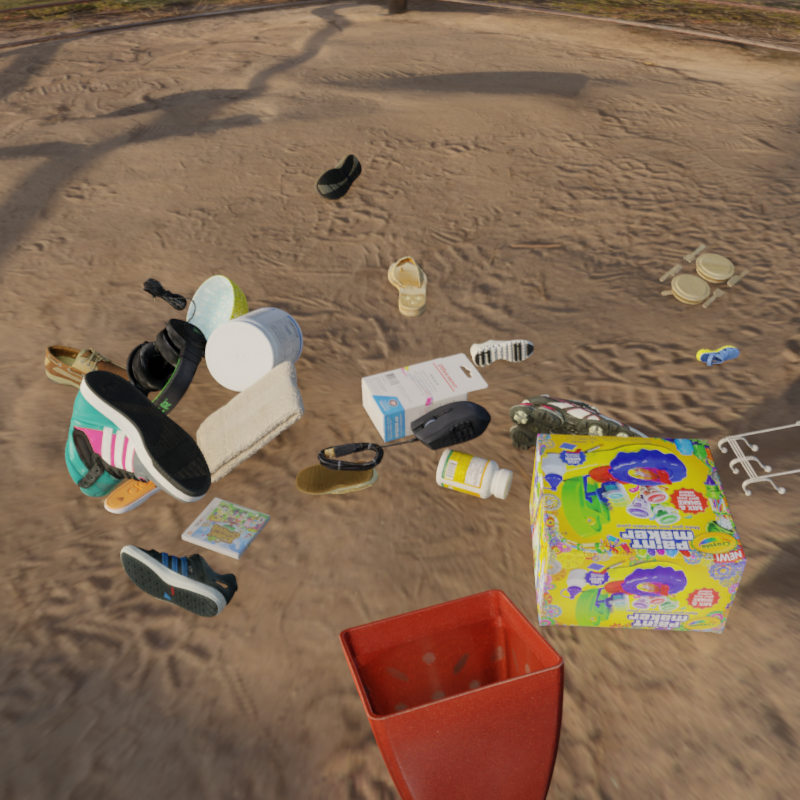} 
        \caption{Input image}
        \label{fig:column_a}
    \end{subfigure}
    \hfill
    \begin{subfigure}[t]{0.32\linewidth}
        \centering
        \includegraphics[width=\textwidth]{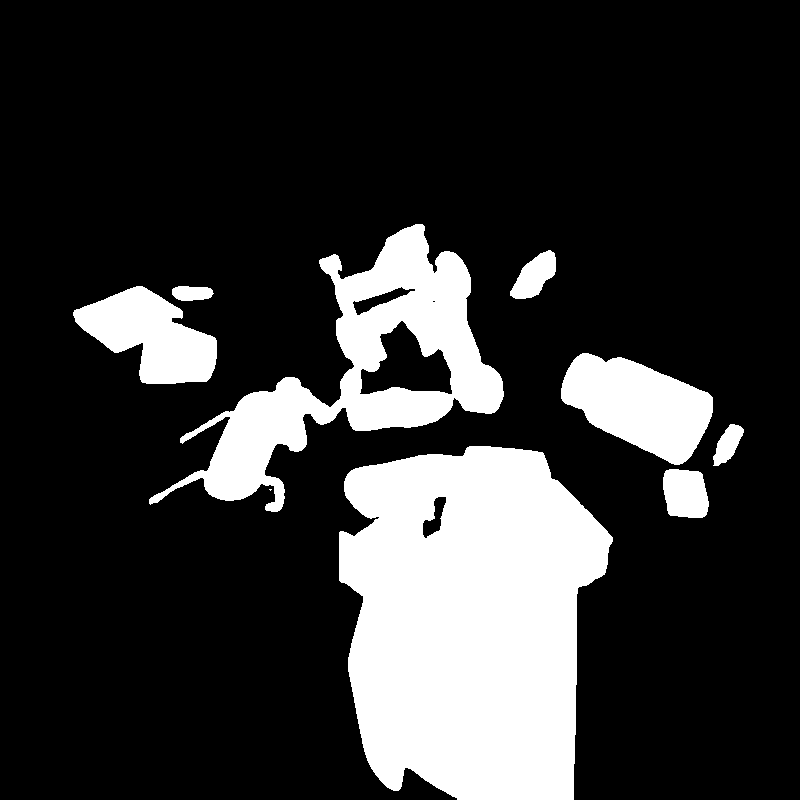} \\ 
        \includegraphics[width=\textwidth]{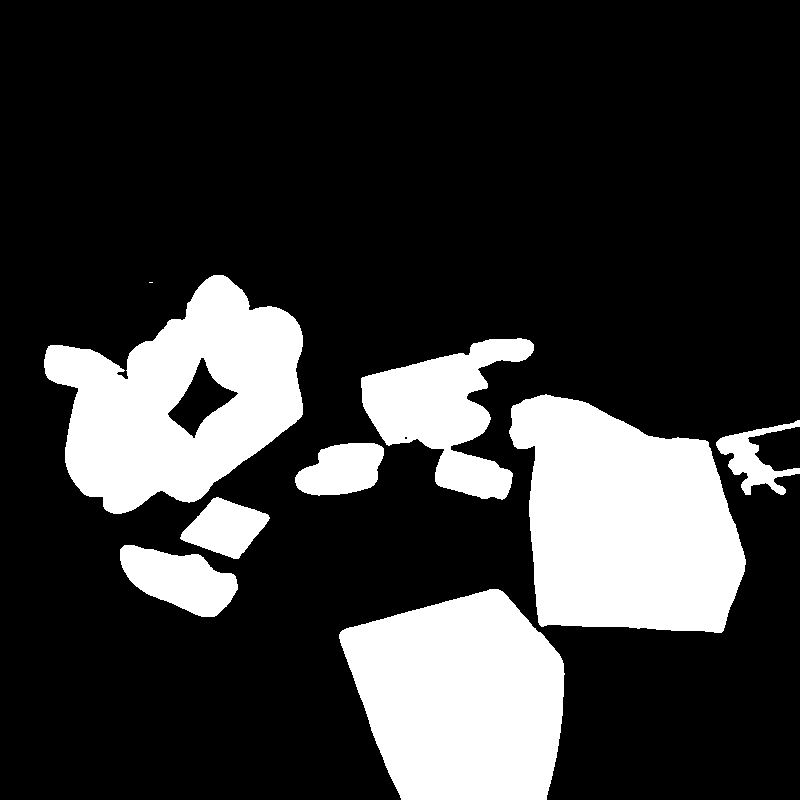} 
        \caption{TAM mask}
        \label{fig:column_b}
    \end{subfigure}
    \hfill
    \begin{subfigure}[t]{0.32\linewidth} 
        \centering
        \includegraphics[width=\textwidth]{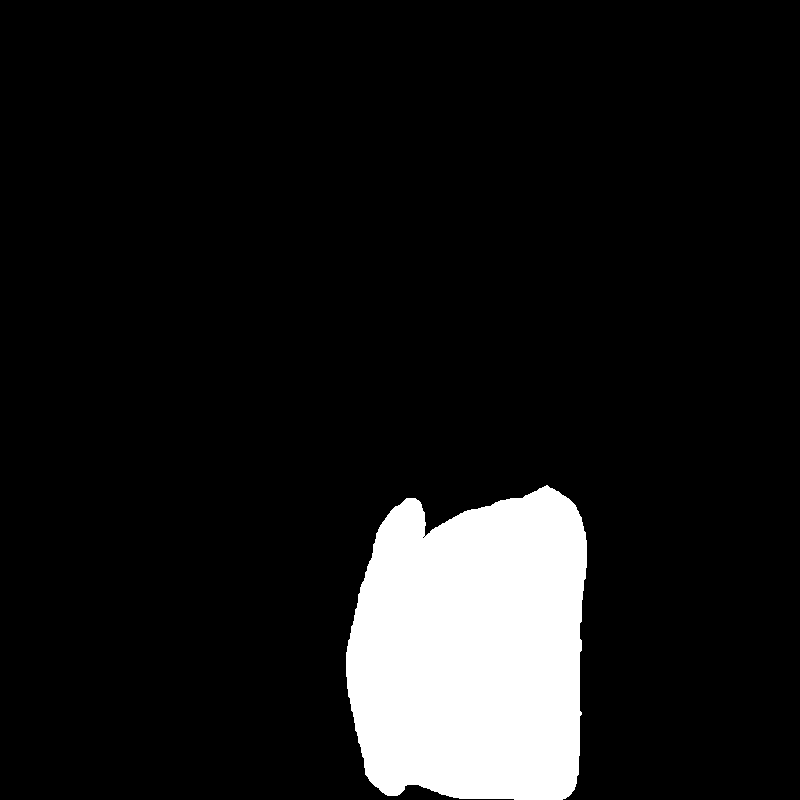} \\   
        \includegraphics[width=\textwidth]{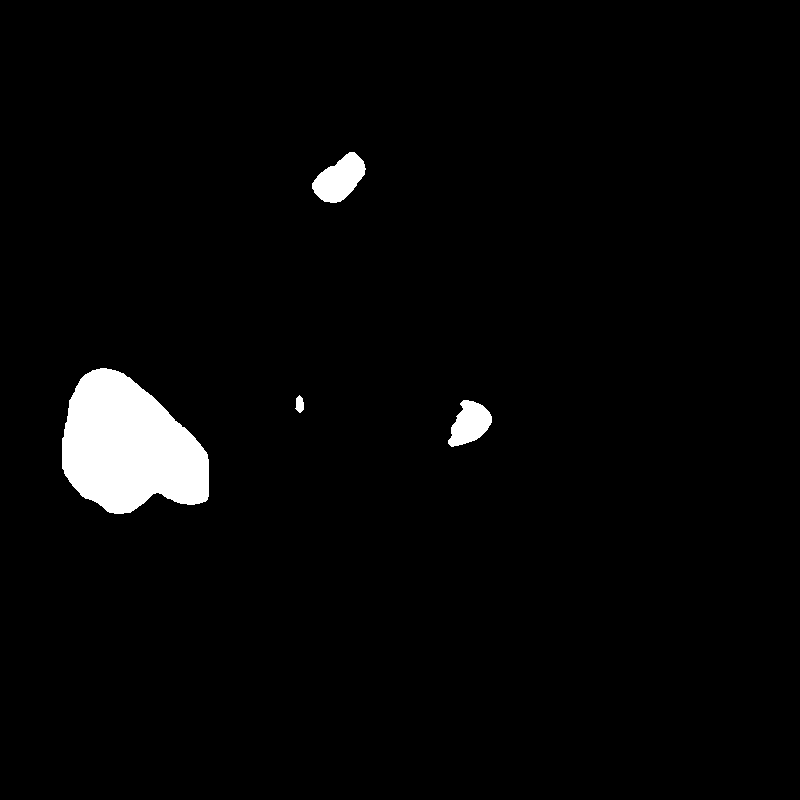}  
        \caption{RoDynRF mask}
        \label{fig:column_c}
    \end{subfigure}
    \vspace{-2mm}
    \caption{\textbf{Comparison of Motion Masks.} TAM~\citep{yang2023track} produces more accurate dynamic object masks compared to those obtained using the masking method of RoDynRF~\cite{liu2023robust}.}
    \vspace{-5mm}
    \label{fig:motion_masks}
\end{figure}

RoDynRF~\citep{liu2023robust} uses motion masks derived from video frames to distinguish static and dynamic elements in scenes.
To achieve this, RoDynRF leverages forward and backward optical flows of the video frames estimated by RAFT~\cite{teed2020raft}.
Then, it uses 8-point and RANSAC algorithm to estimate the fundamental matrix between adjacent frames.
Afterward, it computes the error between the points projected using the fundamental matrix and those derived from the predicted flows.
Regions with high error are assumed to correspond to dynamic parts of the frames.
Although this approach produces reliable masks on previous benchmarks, we observed that this method often fails due to the large motion observed between adjacent frames in Kubric-MRig and iPhone dataset.
Moreover, RoDynRF relies on hard-coded class names to generate clear masks, which involves failure for scenes with significant motion with non-hard-coded instances, thus limiting its practical applicability. RoDyGS instead leverages motion masks generated by the TAM~\cite{yang2023track}, similar to those used in Shape of Motion~\cite{wang2024shape}.
We visualize generated masks in Figure~\ref{fig:motion_masks}.

\subsection{Automatic Motion Masking using SAM}
As mentioned in the previous section, RoDyGS employs TAM to extract motion masks for each frame in casual videos. Alternatively, we investigate the use of Segment Anything Model (SAM)~\cite{ravi2024sam} to automatically mask dynamic objects with a text prompt. After testing various prompts, we found that the prompt `\textit{objects}.' provided the best results.

\begin{figure*}[!htb]
    \centering
    \includegraphics[width=1.0\linewidth]{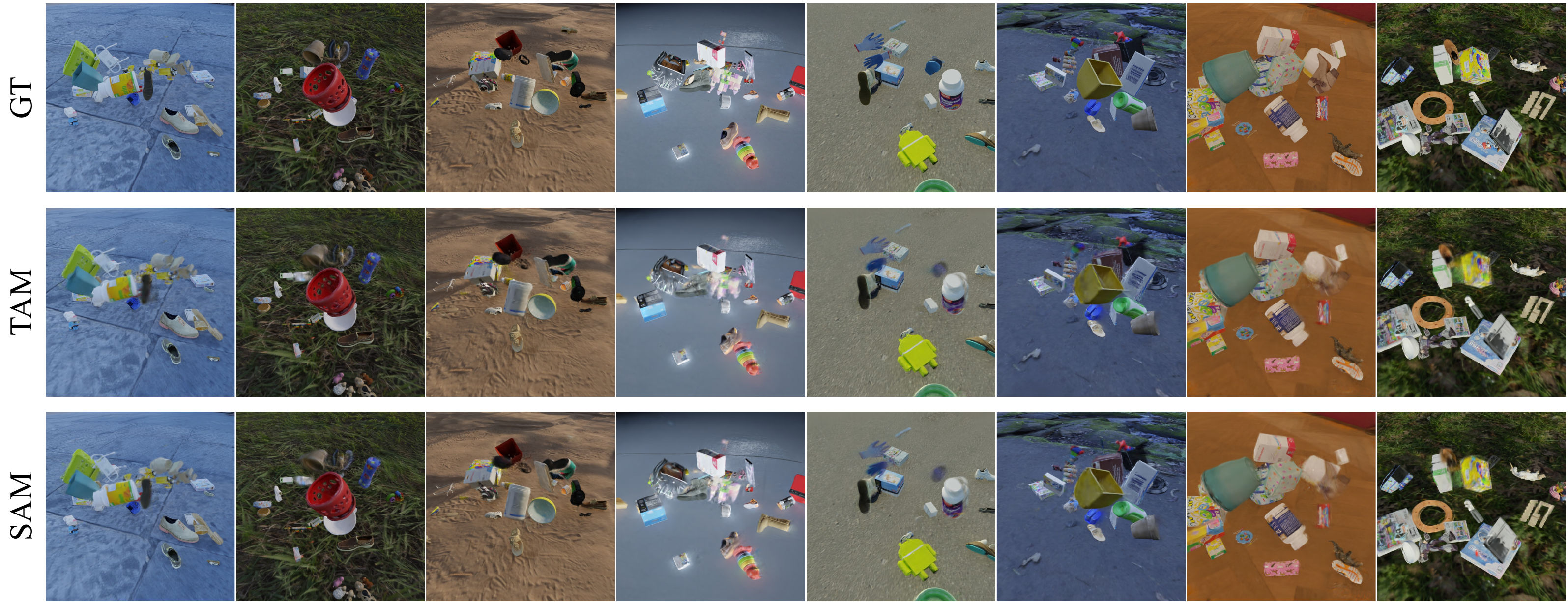}
    \caption{\textbf{
Comparison of RoDyGS when leveraging SAM~\cite{ravi2024sam} and TAM~\cite{yang2023track} on Kubric-MRig dataset.} RoDyGS with motion masks obtained by SAM achieves competitive visual quality to RoDyGS with motions masks obtained by TAM.}
    \label{fig:suppl_sam_tam}
\end{figure*}

Table~\ref{tab:supp_sam_exp} and Figure~\ref{fig:suppl_sam_tam} compare RoDyGS with two variants: one using motion masks obtained by SAM and the other by TAM. Our findings indicate that there is no significant performance gap when using SAM. However, SAM is limited in its ability to extract motion masks in scenes with many stationary objects or dynamic backgrounds. Consequently, we opt to use TAM instead of SAM for improved quality, despite the minor manual efforts required.

\begin{table*}[!htb]
    \centering
    \renewcommand{\arraystretch}{1.1}
    
    \begin{tabular}{l ccc ccc}
        \toprule
        \multirow{2}{*}{Method} & \multicolumn{3}{c}{Kubric-MRig} & \multicolumn{3}{c}{iPhone~\cite{gao2022dycheck}} \\
        \cmidrule(lr){2-4} \cmidrule(lr){5-7}
        & PSNR(↑) & SSIM(↑) & LPIPS(↓) & PSNR(↑) & SSIM(↑) & LPIPS(↓) \\
        \midrule
        RoDynRF~\cite{liu2023robust} & 18.36 & 0.6121 & 0.5984 & 14.91 & 0.4050 & 0.5531 \\ 
        \ \ \ \ \  + TAM~\cite{yang2023track} & 16.93 & 0.6157 & 0.6224 & 13.87 & 0.3670 & 0.5900 \\ 
        \midrule
        RoDyGS (Ours) & \textbf{19.44} & \textbf{0.7169} & \textbf{0.3401} & \textbf{17.38} & \textbf{0.4656} & \textbf{0.4361} \\
        \ \ \ \ \  + SAM~\cite{ravi2024sam} & 19.41 & 0.7152 & 0.3411 & 17.22 & 0.4606 & 0.4417 \\
        \bottomrule
    \end{tabular}
    
    \caption{\textbf{Dynamic novel view synthesis quality comparison on Kubric-MRig and iPhone datasets.} We compare the performance of different motion mask extraction methods (SAM vs. TAM) on both datasets.}
    \label{tab:supp_sam_exp}
\end{table*}

\section{Limitations and Future Work}

\paragraph{Dynamic Part Separation.} To handle in-the-wild videos, we aim to develop a robust optimization method leveraging off-the-shelf models. 
Our current procedure for separating dynamic and static parts relies on user interaction, making the framework partially manual. 
Segmenting dynamic parts in a video is a challenging task and remains an open topic in video processing. 
One future direction is to automate the separation of dynamic components during optimization. Our observations reveal that the static and dynamic parts were not clearly separated, leading to instability in the optimization process. 
A naive, automated dynamic part extraction process would be developed to mitigate this issue.

\vspace{-3mm}
\paragraph{Camera Calibration.} 
From camera calibration perspective, we intend to investigate advanced joint camera optimization techniques. 
Currently, our RoDyGS framework is heavily dependent on the initial camera parameters estimated MASt3R~\cite{leroy2024grounding}. 
In cases where MASt3R provides inaccurate camera parameters, RoDyGS experiences diminished rendering quality. 
To alleviate this dependency, future research should focus on developing innovative methods that can robustly accommodate incorrectly calibrated camera poses. 
Furthermore, future studies should aim to estimate camera intrinsic parameters to enhance the accuracy of scene reconstruction.
\vspace{-3mm}

\paragraph{Potential of 4DGS Representation.} 

Our method effectively converts 2D videos into 4D representations, making it valuable for a wide range of applications. 
Our 4D Gaussian representation captures 3D dynamics alongside the geometry of the scene, providing rich and detailed information that is challenging to infer directly from 2D perception alone.
As pointed out by \cite{jeong2022perfception}, unlike 2D perception, 3D-based perception directly represents real-world scene information, improving efficiency by eliminating the need for additional adaptors to convert 2D data into 3D representations. 
Further enhancing the informativeness of our 4D representation would significantly improve the accuracy and applicability of perception systems in various real-world scenarios.

\section{Additional Baseline Evaluation}

\begin{figure*}[!htb]
    \centering
    \includegraphics[width=1\textwidth]{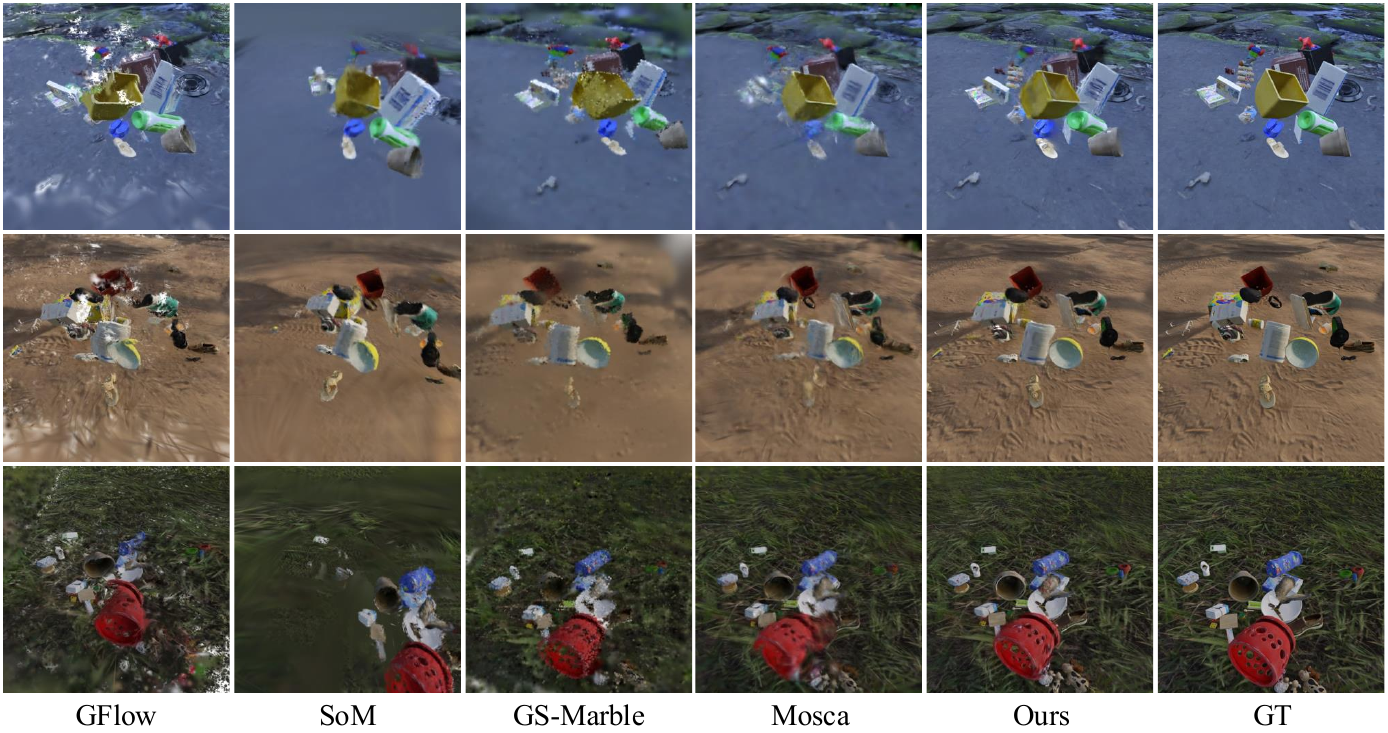}
    \caption{\textbf{Comparison with recent pose-free DVS baselines on Kubric-MRig dataset.} We train all methods using the same MASt3R-initialized camera poses and qualitatively compare their novel view rendering results.}
    \label{fig:recent_baselines_mast3r}
    \clearpage
\end{figure*}

To assess performance on complex dynamic scenes, we benchmark RoDyGS against recent long-term point-tracking methods~\cite{wang2024gflow, wang2024shape, stearns2024dynamic, lei2025mosca}. Each baseline is evaluated under its default settings (Table~\ref{tab:recent_baseline_comparison}) and with MASt3R initialization—using MASt3R camera poses and identical TAM masks (Table~\ref{tab:recent_baseline_comparison_mast3r}). Across both experiments, RoDyGS consistently outperforms the competing methods. We find that GS-Marble fails to converge without MASt3R camera pose priors, indicating sensitivity to large camera motion. Figure~\ref{fig:recent_baselines_mast3r} presents qualitative comparisons for MASt3R-initialized methods: in highly dynamic sequences, all baselines exhibit holes and misalignments from noisy long-term tracks, whereas RoDyGS delivers more realistic novel view renderings.

\begin{table}[!t]
    \centering
    \resizebox{\linewidth}{!}{
    \begin{tabular}{lccc}
        \toprule
         Method & PSNR(↑) & SSIM(↑) & LPIPS(↓) \\
        \midrule
        
         GFlow~\cite{wang2024gflow}       & 16.52 & 0.5770 & 0.5414 \\
        SoM~\cite{som2024}               & 16.83 & 0.6283 & 0.6682 \\
     GS-Marble~\cite{stearns2024dynamic} & - & - & - \\
         Mosca~\cite{lei2025mosca}        & 17.11 & 0.6099 & 0.5547 \\
         
        \rowcolor{gray!20} 
         Ours        & \textbf{19.44} & \textbf{0.7169} & \textbf{0.3401} \\
        \bottomrule
        \end{tabular}
        }
    \vspace{-2mm}
    \caption{\textbf{Novel view synthesis results on the Kubric-MRig dataset.} \emph{GS-Marble} failed to converge in some scenes.
}
    \label{tab:recent_baseline_comparison}
    \vspace{-2mm}
\end{table}

\begin{table}[!t]
    \centering
    \resizebox{\linewidth}{!}{
    \begin{tabular}{lccc}
        \toprule
         Method & PSNR(↑) & SSIM(↑) & LPIPS(↓) \\
        \midrule
        
        GFlow~\cite{wang2024gflow}       & 16.52 & 0.5770 & 0.5414 \\
        SoM~\cite{som2024}               & 16.97 & 0.6232 & 0.5523 \\
        GS-Marble~\cite{stearns2024dynamic} & 15.09 & 0.5743 & 0.6551 \\
        Mosca~\cite{lei2025mosca}        & 19.30 & 0.6642 & 0.4782 \\
        \rowcolor{gray!20} 
         Ours        & \textbf{19.44} & \textbf{0.7169} & \textbf{0.3401} \\
        \bottomrule
        \end{tabular}
        }
    \vspace{-2mm}
    \caption{\textbf{Novel view synthesis results on the Kubric-MRig dataset with MASt3R initialization.} All models were initialized using camera poses estimated by MASt3R~\cite{leroy2024grounding}.}
    \label{tab:recent_baseline_comparison_mast3r}
    \vspace{-2mm}
\end{table}

\section{Evaluation on NVIDIA Dynamic Dataset}

\begin{figure*}[!htb]
    \centering
    \includegraphics[width=1.0\linewidth]{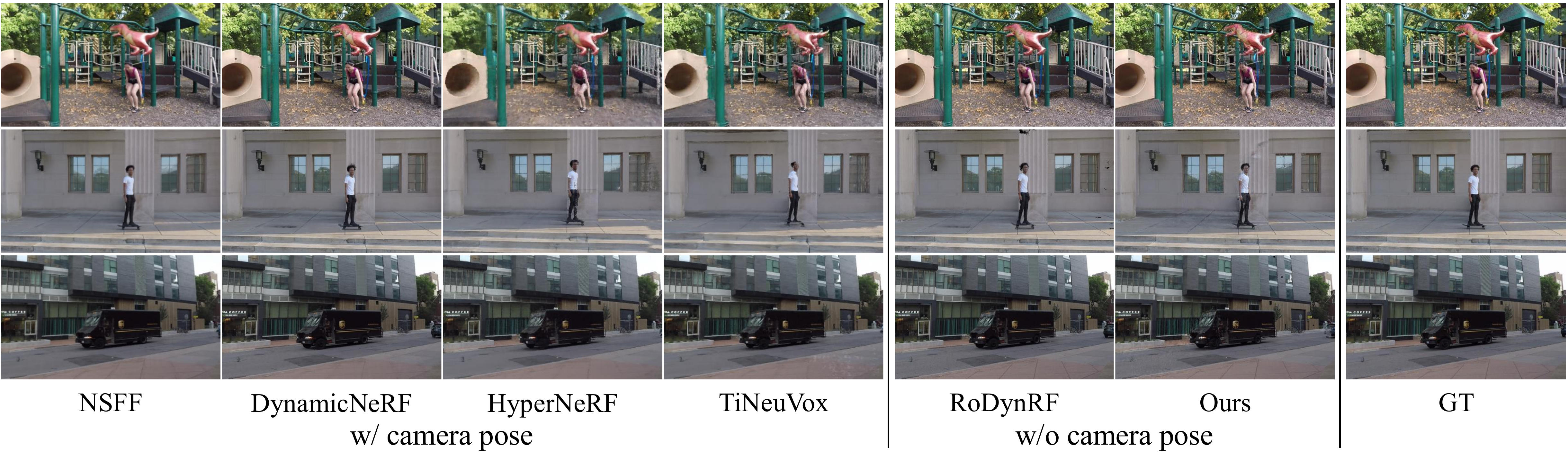}
    \caption{\textbf{Dynamic novel view synthesis on NVIDIA Dynamic dataset.} We compare RoDyGS with RoDynRF on NVIDIA Dynamic with the pose-free setup. RoDyGS synthesizes realistic images similar to those of RoDynRF.}
    \label{fig:suppl_nvidia}
\end{figure*}

\begin{table*}[!ht]
\vspace{-2mm}
\centering
\resizebox{1\linewidth}{!} 
{
\begin{tabular}{lc| ccccccc|c}
\toprule 
PSNR ($\uparrow$) / LPIPS ($\downarrow$) & GT Cam& Jumping & Skating & Truck & Umbrella & Balloon1 & Balloon2 & Playground & Average \\
\midrule
DynamicNeRF*~\cite{Gao-ICCV-DynNeRF} & \checkmark &
24.68 / 0.090 & 
\textbf{32.66} / \textbf{0.035} & 
28.56 / 0.082 & 
23.26 / 0.137 &
22.36 / 0.104 & 
\textbf{27.06} / \textbf{0.049} & 
24.15 / 0.080 &
\textbf{26.10} / 0.082 \\
HyperNeRF~\cite{park2021hypernerf} & \checkmark &
18.34 / 0.302 &
21.97 / 0.183 & 
20.61 / 0.205 & 
18.59 / 0.443 & 
13.96 / 0.530 & 
16.57 / 0.411 & 
13.17 / 0.495 &
17.60 / 0.367 \\
TiNeuVox~\cite{TiNeuVox} & \checkmark &
20.81 / 0.247 &
23.32 / 0.152 & 
23.86 / 0.173 & 
20.00 / 0.355 & 
17.30 / 0.353 & 
19.06 / 0.279 & 
13.84 / 0.437 &
19.74 / 0.285 \\
RoDynRF~\cite{liu2023robust} &  \checkmark&
\textbf{25.66} / \textbf{0.071} & 
28.68 / 0.040 & 
\textbf{29.13} / \textbf{0.063} & 
\textbf{24.26} / \textbf{0.089} & 
\textbf{22.37} / \textbf{0.103} & 
26.19 / 0.054 & 
\textbf{24.96} / \textbf{0.048} & 
25.89 / \textbf{0.065} \\
\midrule
RoDynRF~\cite{liu2023robust} & - &
24.27 / 0.100 & 
28.71 / 0.046 & 
28.85 / 0.066 & 
23.25 / 0.104 & 
21.81 / 0.122 & 
25.58 / 0.064 & 
25.20 / 0.052 &
25.38 / 0.079 \\
RoDyGS (Ours) &  -&
22.86 / 0.214 &
26.30 / 0.089 &
26.58 / 0.134 &
22.93 / 0.170 &
20.94 / 0.224 &
25.22 / 0.128 &
23.30 / 0.095 &
24.02 / 0.151 \\
\bottomrule
\end{tabular}
}
\caption{
\textbf{Dynamic novel view synthesis results on NVIDIA Dynamic dataset.}
Comparision of average PSNR and LPIPS evaluation results to existing methods on NVIDIA Dynamic dataset~\cite{yoon2020novel}.
}
\label{tab:quantitative_nvidia}
\end{table*}
We evaluate RoDyGS against several baseline methods on NVIDIA Dynamic dataset~\cite{yoon2020novel}. 
This dataset comprises 7 scenes, each captured with 12 fixed cameras arranged in a forward-facing configuration. 
Table~\ref{tab:quantitative_nvidia} presents PSNR and LPIPS scores obtained from NVIDIA Dynamic dataset. 
In Figure~\ref{fig:suppl_nvidia} visualizes the rendered results on NVIDIA Dynamic dataset. 

We experimentally observed that RoDynRF outperforms RoDyGS on NVIDIA Dynamic dataset.
We hypothesize that this phenomenon occurs because RoDynRF leverages normalized device coordinates~\cite{mildenhall2019llff}, which assume that all scenes are captured in a forward-facing manner~\cite{yoon2020novel}, providing a significant advantage over our method.
Additionally, the dataset includes ambient camera motion and lacks temporal consistency across adjacent frames, making it less representative of real-world scenarios.

\section{Scene-wise Experimental Results}

\paragraph{Quantitative results.}
We report scene-wise scores to facilitate future research. 
Table~\ref{tab:psnr_kubric_mrig_all}, \ref{tab:ssim_kubric_mrig_all}, \ref{tab:lpips_kubric_mrig_all}, \ref{tab:ate_kubric_mrig_all}, \ref{tab:rpe_t_kubric_mrig_all}, and \ref{tab:rpe_r_kubric_mrig_all} report scene-wise PSNR, SSIM, LPIPS, ATE, RPE-t, and RPE-R scores on Kubric-MRig dataset, respectively. 
Then, Table~\ref{tab:psnr_iphone_all_results}, \ref{tab:ssim_iphone_all_results}, \ref{tab:lpips_iphone_all_results}, \ref{tab:ate_iphone_all}, \ref{tab:rpe_t_iphone_all}, and \ref{tab:rpe_r_iphone_all} report scene-wise PSNR, SSIM, LPIPS, ATE, RPE-t, and RPE-R scores on the iPhone~\cite{gao2022dycheck} dataset, respectively. "GT Cam" in tables refers to the availability of ground truth camera poses during training. 

\begin{figure*}[!htb]
    \centering
    \includegraphics[width=1.0\linewidth]{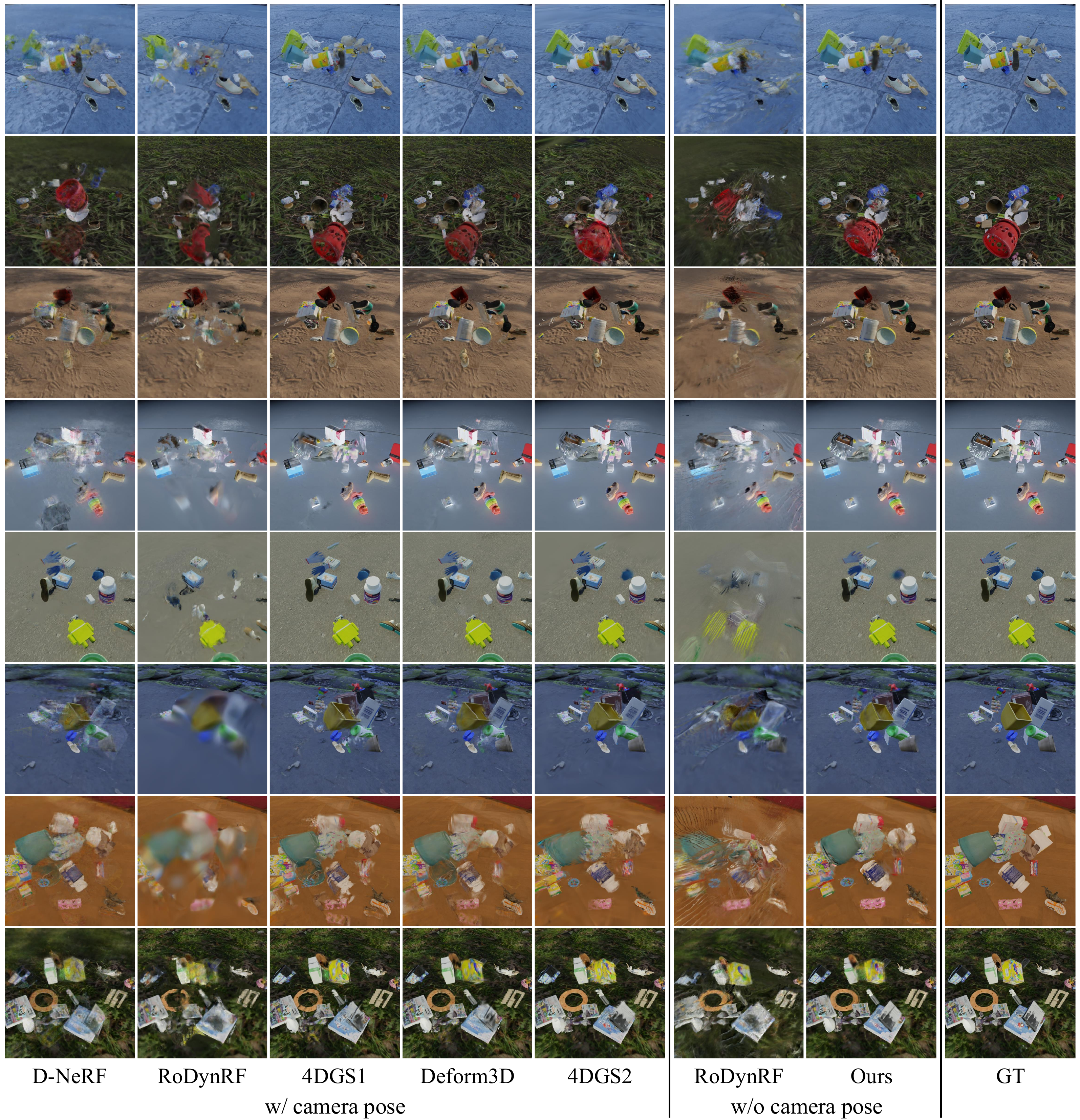}
    \caption{\textbf{Dynamic novel view synthesis on Kubric-MRig dataset.} Comparison of rendering results between RoDyGS and other dynamic neural field methods, both pose-aware~\cite{pumarola2020d, liu2023robust, wu20244d, yang2023gs4d, yang2024deformable} and pose-free~\cite{liu2023robust}. In the pose-free setup, RoDyGS produces clearer rendering results than RoDynRF.}
    \label{fig:suppl_kubric}
\end{figure*}

\begin{figure*}[!htb]
    \centering
    \includegraphics[width=1.0\linewidth]{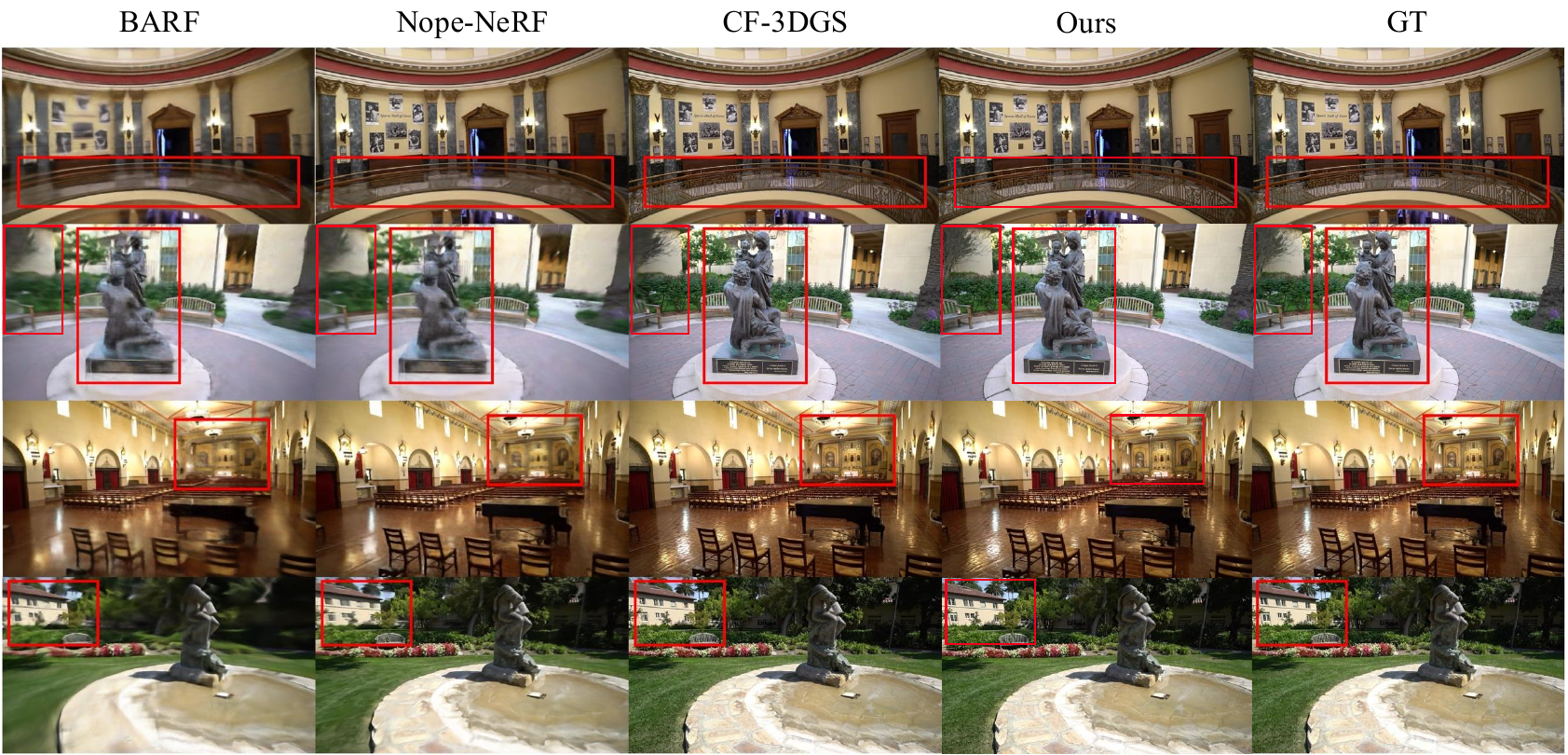}
    \caption{\textbf{Static novel view synthesis on Tanks and Temples dataset.} Comparison on the Tanks and Temples dataset between RoDyGS and previous pose-free neural field methods ~\cite{lin2021barf, bian2023nope, fu2024cf3dgs}. RoDyGS demonstrates competitive rendering quality with CF-3DGS~\cite{fu2024cf3dgs}, the previous state-of-the-art pose-free neural field for static scenes.}
    \label{fig:suppl_tnt}
\end{figure*}

\begin{figure*}[!htb]
    \centering
    \includegraphics[width=1.0\linewidth]{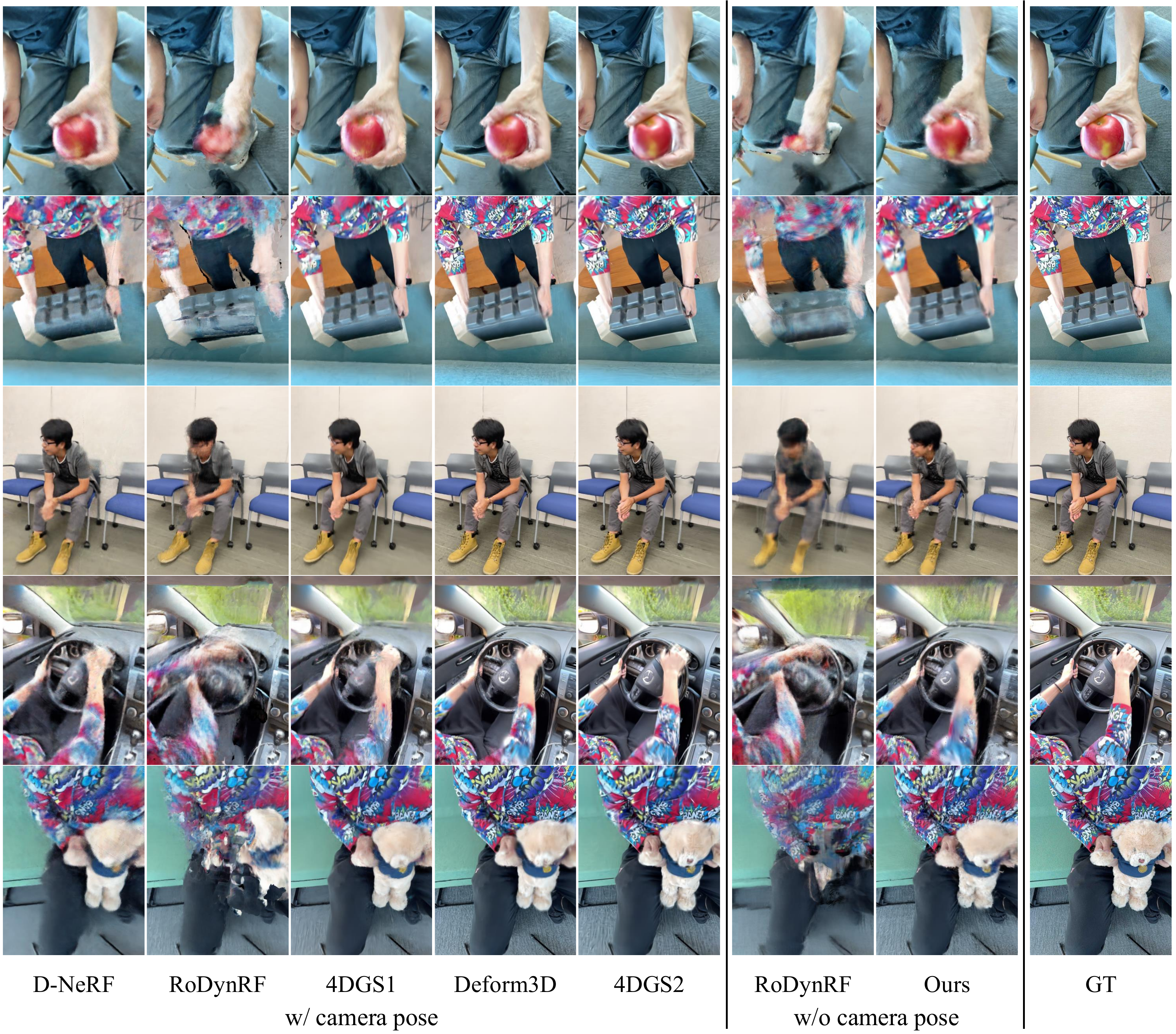}
    \caption{\textbf{Dynamic novel view synthesis on iPhone dataset.} We compare our RoDyGS method against both pose-aware~\cite{pumarola2020d, liu2023robust, wu20244d, yang2023gs4d, yang2024deformable} and pose-free~\cite{liu2023robust} dynamic neural fields. RoDyGS achieves better visual clarity than RoDynRF under the pose-free setup.}
    \label{fig:suppl_iphone}
\end{figure*}

Additonally, we train RoDynRF with TAM masks~\cite{yang2023track} and MASt3R~\cite{leroy2024grounding}, which are the same off-the-shelf priors used in RoDyGS. Table~\ref{tab:ablation_scene_wise} reports the scene-wise NVS results on iPhone dataset. Note that RoDynRF fails to converge on the "block" scene during training when initialized with MASt3R camera poses. 
\paragraph{Qualitative results.}
We visualize additional qualitative results of our work. Figure~\ref{fig:suppl_kubric} visualizes rendered results of baselines and RoDyGS on Kubric-MRig. Then, Figure~\ref{fig:suppl_tnt} presents the comparison between ours and pose-free 3D neural fields on Tanks and Temples. Finally, Figure~\ref{fig:suppl_iphone} demonstrates our method's performance on the iPhone dataset compared to both pose-aware and pose-free approaches. 
\section{Scene-wise Quantitative Results}
We report scene-wise scores to facilitate future research. 
Table~\ref{tab:psnr_kubric_mrig_all}, \ref{tab:ssim_kubric_mrig_all}, \ref{tab:lpips_kubric_mrig_all}, \ref{tab:ate_kubric_mrig_all}, \ref{tab:rpe_t_kubric_mrig_all}, and \ref{tab:rpe_r_kubric_mrig_all} report scene-wise PSNR, SSIM, LPIPS, ATE, RPE-t, and RPE-R scores on Kubric-MRig dataset, respectively. 
Then, Table~\ref{tab:psnr_iphone_all_results}, \ref{tab:ssim_iphone_all_results}, \ref{tab:lpips_iphone_all_results}, \ref{tab:ate_iphone_all}, \ref{tab:rpe_t_iphone_all}, and \ref{tab:rpe_r_iphone_all} report scene-wise PSNR, SSIM, LPIPS, ATE, RPE-t, and RPE-R scores on the iPhone~\cite{gao2022dycheck} dataset, respectively. Note that GT Cam refers to the availability of ground truth camera poses during training.

Additonally, we train RoDynRF with TAM masks~\cite{yang2023track} and MASt3R~\cite{leroy2024grounding}, which are the same off-the-shelf priors used in RoDyGS. Table~\ref{tab:ablation_scene_wise} reports the scene-wise NVS results on iPhone dataset. Note that RoDynRF fails to converge on the "block" scene during training when initialized with MASt3R camera poses.

\begin{table*}[!h]
    \centering
    \begin{tabular}{l ccc ccc ccc}
        \toprule
        \multirow{2}{*}{iPhone~\cite{gao2022dycheck}} & \multicolumn{3}{c}{RoDynRF + TAM~\cite{yang2023track}} & \multicolumn{3}{c}{RoDynRF + MASt3R~\cite{leroy2024grounding}} & \multicolumn{3}{c}{RoDyGS (Ours)} \\
        \cmidrule(lr){2-4} \cmidrule(lr){5-7} \cmidrule(lr){8-10} 
        & PSNR(↑) & SSIM(↑) & LPIPS(↓) & PSNR(↑) & SSIM(↑) & LPIPS(↓) & PSNR(↑) & SSIM(↑) & LPIPS(↓) \\
        \midrule
        apple & 12.99 & 0.3873 & 0.6410 & 16.51 & 0.4260 & 0.5523 & 16.79 & 0.4347 & 0.4850 \\
        block & 14.17 & 0.4504 & 0.6282 & 4.44 & 0.0030 & 0.7533 & 17.67 & 0.5369 & 0.4748 \\
        paper-windmill & 16.18 & 0.2966 & 0.5024 & 14.78 & 0.2184 & 0.5027 & 19.20 & 0.4627 & 0.3569 \\
        space-out & 16.84 & 0.5500 & 0.5299 & 16.06 & 0.5188 & 0.5541 & 19.16 & 0.5808 & 0.4086 \\
        spin & 13.80 & 0.2976 & 0.5793 & 14.55 & 0.2960 & 0.5751 & 18.47 & 0.3921 & 0.4278 \\
        wheel & 11.77 & 0.3014 & 0.5729 & 14.73 & 0.3526 & 0.6912 & 15.66 & 0.4502 & 0.4203 \\
        teddy & 11.35 & 0.2879 & 0.6744 & 15.79 & 0.3777 & 0.6036 & 14.69 & 0.4015 & 0.4790 \\
        \bottomrule  
    \end{tabular}
    
    \caption{\textbf{Scene-wise comparison between RoDynRF with off-the-shelf priors and RoDyGS.} We compare the dynamic novel view synthesis quality on the iPhone dataset between RoDyGS, RoDynRF initialized with MASt3R~\cite{leroy2024grounding} camera poses, and RoDynRF with TAM~\cite{yang2023track} masks.}
    \label{tab:ablation_scene_wise}
\end{table*}

\begin{figure*}[!htb]
    \centering
    \includegraphics[width=1.0\linewidth]{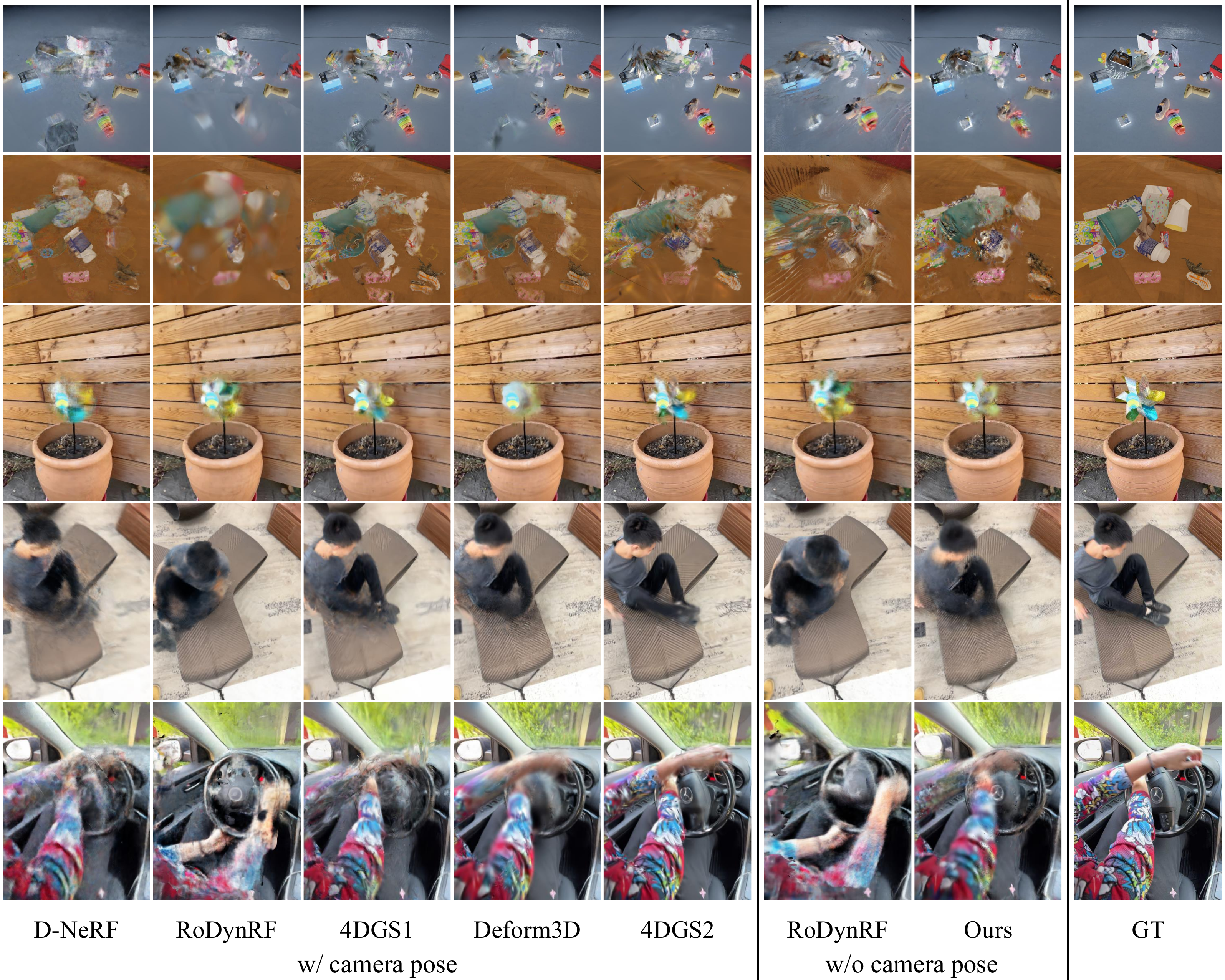}
    \caption{\textbf{Failure cases of RoDyGS.} 
RoDyGs and other baselines struggle from large motions and occlusion in scenes. 
}
    \label{fig:suppl_failure_case}
\end{figure*}

\clearpage

\begin{table*}[!t]
    \centering
    \vspace{-2mm}
    \resizebox{1.0\linewidth}{!}{
    \begin{tabular}{lc|cccccccc|c}
    \toprule
     PSNR($\uparrow$) & GT Cam & scene0 & scene1 & scene2 & scene3 & scene4 & scene5 & scene6 & scene7 & average \\
        \midrule
        D-NeRF~\cite{pumarola2020d} & \checkmark & 21.95 & 19.90 & 22.73 & 20.56 & 23.62 & 20.61 & 19.11 & 16.68 & 20.65 \\
        RoDynRF~\cite{liu2023robust} & \checkmark & 23.19 & 21.44 & 22.82 & 20.56 & 22.39 & 20.74 & 18.00 & 17.30 & 20.80 \\
        4DGS1~\cite{wu20244d} &\checkmark & 24.68 & 23.38 & 24.95 & 21.76 & 23.92 & 22.65 & 20.19 & 18.13 & 22.46 \\
        Deform3D~\cite{yang2024deformable} &\checkmark & 24.31 & 23.57 & 25.51 & 22.12 & 24.15 & 22.44 & 20.68 & 18.50 & 22.66 \\
        4DGS2~\cite{yang2023gs4d} & \checkmark &25.29 & 24.03 & 24.86 & 21.99 & 23.99 & 23.00 & 22.07 & 18.89 & 23.02 \\
        \midrule
        RoDynRF ~\cite{liu2023robust} & - & 20.14 & 18.88 & 18.73 & 17.16 & 20.65 & 19.18 & 17.53 & 14.62 & 18.36 \\
        \ \ \ \ \ + TAM~\cite{yang2023track}  & -& 18.82 & 15.89 & 18.09 & 16.55 & 20.01 & 16.80 & 16.28 & 12.98 & 16.93 \\
        \midrule
        
        RoDyGS (Ours) & - &22.31 & 21.10 & 20.17 & 18.59 & 21.19 & 20.44 & 18.61 & 13.11 & 19.44
 \\
    \bottomrule
    \end{tabular}
    }
    \vspace{-2mm}
    \caption{\textbf{
PSNR results for Kubric-MRig dataset.}}
    \label{tab:psnr_kubric_mrig_all}
\end{table*}
\begin{table*}[!t]
    \centering
    \vspace{-2mm}
    \resizebox{1.0\linewidth}{!}{
    \begin{tabular}{lc|cccccccc|c}
    \toprule
     SSIM($\uparrow$) & GT Cam & scene0 & scene1 & scene2 & scene3 & scene4 & scene5 & scene6 & scene7 & average \\
        \midrule
        D-NeRF~\cite{pumarola2020d} & \checkmark & 0.7359 & 0.5533 & 0.7468 & 0.8069 & 0.7942 & 0.7299 & 0.7409 & 0.6202 & 0.7160 \\
        RoDynRF~\cite{liu2023robust} & \checkmark & 0.7773 & 0.6935 & 0.7854 & 0.8361 & 0.7299 & 0.8146 & 0.7707 & 0.6594 & 0.7584 \\
        4DGS1~\cite{wu20244d} & \checkmark & 0.8967 & 0.8802 & 0.9035 & 0.8563 & 0.8913 & 0.8639 & 0.7614 & 0.7924 & 0.8557 \\
        Deform3D~\cite{yang2024deformable} & \checkmark & 0.8178 & 0.8791 & 0.8998 & 0.8675 & 0.8857 & 0.8513 & 0.7803 & 0.8013 & 0.8478 \\
        4DGS2~\cite{yang2023gs4d} & \checkmark & 0.8695 & 0.8623 & 0.8846 & 0.8467 & 0.8806 & 0.8528 & 0.7856 & 0.7861 & 0.8460 \\
        \midrule
        RoDynRF~\cite{liu2023robust} & - & 0.6255 & 0.4772 & 0.6295 & 0.7153 & 0.6728 & 0.6958 & 0.6379 & 0.4432 & 0.6121 \\
        \ \ \ \ \ + TAM~\cite{yang2023track} & - & 0.6245 & 0.4457 & 0.6332 & 0.7351 & 0.6965 & 0.6522 & 0.6994 & 0.4393 & 0.6157 \\
        \midrule
        RoDyGS (Ours) & - & 0.7549 & 0.7043 & 0.7383 & 0.7667 & 0.7596 & 0.7810 & 0.6943 & 0.5360 & 0.7169 \\
        \bottomrule
    \end{tabular}
    }
    \vspace{-2mm}
    \caption{\textbf{SSIM results for Kubric-MRig dataset.}}
    \label{tab:ssim_kubric_mrig_all}
\end{table*}
\begin{table*}[!t]
    \centering
    \vspace{-2mm}
    \resizebox{1.0\linewidth}{!}{
    \begin{tabular}{lc|cccccccc|c}
    \toprule
     LPIPS ($\downarrow$) & GT Cam & scene0 & scene1 & scene2 & scene3 & scene4 & scene5 & scene6 & scene7 & average \\
        \midrule
        D-NeRF~\cite{pumarola2020d} & \checkmark & 0.4165 & 0.5112 & 0.3884 & 0.3357 & 0.3521 & 0.4543 & 0.3222 & 0.4364 & 0.4021 \\
        RoDynRF~\cite{liu2023robust} & \checkmark & 0.4443 & 0.4627 & 0.4244 & 0.4427 & 0.5494 & 0.5074 & 0.5936 & 0.4438 & 0.4836 \\
        4DGS1~\cite{wu20244d} & \checkmark & 0.1192 & 0.1279 & 0.1122 & 0.2054 & 0.1339 & 0.1935 & 0.3090 & 0.2073 & 0.1760 \\
        Deform3D~\cite{yang2024deformable} & \checkmark & 0.2013 & 0.1301 & 0.1300 & 0.1989 & 0.1445 & 0.2000 & 0.2974 & 0.1990 & 0.1877 \\
        4DGS2~\cite{yang2023gs4d} & \checkmark & 0.1563 & 0.1455 & 0.1407 & 0.2263 & 0.1457 & 0.2012 & 0.2860 & 0.2168 & 0.1898 \\
        \midrule
        RoDynRF~\cite{liu2023robust} & - & 0.5869 & 0.6555 & 0.5811 & 0.5357 & 0.5883 & 0.6062 & 0.6258 & 0.6081 & 0.5984 \\
        \ \ \ \ \ + TAM~\cite{yang2023track} & - & 0.6246 & 0.7292 & 0.6203 & 0.5343 & 0.5917 & 0.6120 & 0.6200 & 0.6472 & 0.6224 \\
        \midrule
        RoDyGS (Ours) & - & 0.2928 & 0.2818 & 0.3130 & 0.3413 & 0.2938 & 0.3149 & 0.4347 & 0.4481 & 0.3401 \\
        \bottomrule
    \end{tabular}
    }
    \vspace{-2mm}
    \caption{\textbf{LPIPS results for Kubric-MRig dataset.}}
    \label{tab:lpips_kubric_mrig_all}
\end{table*}

\begin{table*}[!t]
    \centering
    \vspace{-2mm}
    \resizebox{1.0\linewidth}{!}{
    \begin{tabular}{l|cccccccc|c}
    \toprule
     ATE($\downarrow$) & scene0 & scene1 & scene2 & scene3 & scene4 & scene5 & scene6 & scene7 & average \\
        \midrule
        RoDynRF~\cite{liu2023robust} & 0.0881 & 0.0468 & 0.0824 & 0.0724 & 0.0319 & 0.0603 & 0.0446 & 0.0793 & 0.0632 \\
        \ \ \ \ \ + TAM~\cite{yang2023track} & 0.0818 & 0.0693 & 0.0826 & 0.0764 & 0.0361 & 0.0801 & 0.0704 & 0.0593 & 0.0695 \\
        \midrule
        RoDyGS (Ours) & 0.0009 & 0.0025 & 0.0071 & 0.0061 & 0.0045 & 0.0088 & 0.0028 & 0.0085 & 0.0052 \\
        \bottomrule
    \end{tabular}
    }
    \vspace{-2mm}
    \caption{\textbf{
ATE results for Kubric-MRig dataset.}}
    \label{tab:ate_kubric_mrig_all}
\end{table*}

\begin{table*}[!t]
    \centering
    \vspace{-2mm}
    \resizebox{1.0\linewidth}{!}{
    \begin{tabular}{l|cccccccc|c}
    \toprule
     RPE-t($\downarrow$) & scene0 & scene1 & scene2 & scene3 & scene4 & scene5 & scene6 & scene7 & average \\
        \midrule
        RoDynRF~\cite{liu2023robust} & 0.5425 & 0.3475 & 0.5030 & 0.4713 & 0.2174 & 0.3328 & 0.3190 & 0.5368 & 0.4088 \\
        \ \ \ \ \ + TAM~\cite{yang2023track}& 0.7732 & 0.4195 & 0.4228 & 0.6579 & 0.2550 & 0.4593 & 0.4750 & 0.2982 & 0.4701 \\
        \midrule
        RoDyGS (Ours) & 0.0083 & 0.0207 & 0.0563 & 0.0512 & 0.0320 & 0.0479 & 0.0223 & 0.0601 & 0.0374 \\
        \bottomrule
    \end{tabular}
    }
    \vspace{-2mm}
    \caption{\textbf{
RPE-t results for Kubric-MRig dataset.}}
    \label{tab:rpe_t_kubric_mrig_all}
\end{table*}

\begin{table*}[!t]
    \centering
    \vspace{-2mm}
    \resizebox{1.0\linewidth}{!}{
    \begin{tabular}{l|cccccccc|c}
    \toprule
     RPE-R($\downarrow$) & scene0 & scene1 & scene2 & scene3 & scene4 & scene5 & scene6 & scene7 & average \\
        \midrule
        RoDynRF~\cite{liu2023robust} & 1.9904 & 1.6781 & 2.1122 & 1.7471 & 1.5009 & 1.9715 & 1.7836 & 1.8201 & 1.8255 \\
        \ \ \ \ \ + TAM~\cite{yang2023track} & 2.2493 & 1.7614 & 2.4783 & 1.8486 & 1.5099 & 1.8526 & 1.8043 & 1.4920 & 1.8746 \\
        \midrule
        RoDyGS (Ours) & 0.0187 & 0.0739 & 0.1004 & 0.1170 & 0.0902 & 0.1494 & 0.1036 & 0.0933 & 0.0933 \\
        \bottomrule
    \end{tabular}
    }
    \vspace{-2mm}
    \caption{\textbf{
RPE-R results for Kubric-MRig dataset.}}
    \label{tab:rpe_r_kubric_mrig_all}
\end{table*}

\begin{table*}[!t]
    \centering
    \vspace{-2mm}
    \resizebox{1.0\linewidth}{!}{
    \begin{tabular}{lc|ccccccc|c}
    \toprule
     PSNR($\uparrow$) & GT Cam & apple & block & paper-windmill & space-out & spin & teddy & wheel & average \\
        \midrule
        D-NeRF~\cite{pumarola2020d} & \checkmark & 24.23 & 21.80 & 21.85 & 25.18 & 22.15 & 19.46 & 19.65 & 22.04 \\
        RoDynRF~\cite{liu2023robust} & \checkmark & 17.38 & 15.99 & 20.71 & 20.62 & 16.66 & 13.28 & 12.78 & 16.78 \\
        4DGS1~\cite{wu20244d} & \checkmark & 23.24 & 22.05 & 21.03 & 24.81 & 22.99 & 18.89 & 17.72 & 21.25 \\
        Deform3D~\cite{yang2024deformable} & \checkmark & 24.82 & 23.26 & 20.62 & 26.95 & 23.51 & 20.93 & 19.56 & 22.81 \\
        4DGS2~\cite{yang2023gs4d} & \checkmark & 27.31 & 25.09 & 26.81 & 29.54 & 26.30 & 21.11 & 22.45 & 25.52 \\
        \midrule
        RoDynRF~\cite{liu2023robust} & - & 14.50 & 14.73 & 17.94 & 18.06 & 15.75 & 11.56 & 11.85 & 14.91 \\
        \ \ \ \ \ + TAM~\cite{yang2023track} & - & 12.99 & 14.17 & 16.18 & 16.84 & 13.80 & 11.35 & 11.77 & 13.87 \\
        \midrule
        RoDyGS (Ours) & - & 16.79 & 17.67 & 19.20 & 19.16 & 18.47 & 14.69 & 15.66 & 17.38 \\
    \bottomrule
    \end{tabular}
    }
    \vspace{-2mm}
    \caption{\textbf{PSNR results for iPhone dataset.}}
    \label{tab:psnr_iphone_all_results}
\end{table*}

\begin{table*}[!t]
    \centering
    \vspace{-2mm}
    \resizebox{1.0\linewidth}{!}{
    \begin{tabular}{lc|ccccccc|c}
    \toprule
     SSIM($\uparrow$) & GT Cam & apple & block & paper-windmill & space-out & spin & teddy & wheel & average \\
        \midrule
        D-NeRF~\cite{pumarola2020d} & \checkmark & 0.6748 & 0.6238 & 0.5129 & 0.7018 & 0.5002 & 0.4940 & 0.5665 & 0.5820 \\
        RoDynRF~\cite{liu2023robust} & \checkmark & 0.5310 & 0.4858 & 0.5760 & 0.6844 & 0.5139 & 0.3806 & 0.3959 & 0.5096 \\
        4DGS1~\cite{wu20244d} & \checkmark & 0.6512 & 0.6654 & 0.5080 & 0.7385 & 0.5283 & 0.5562 & 0.5224 & 0.5865 \\
        Deform3D~\cite{yang2024deformable} & \checkmark & 0.7394 & 0.7377 & 0.5557 & 0.8177 & 0.6528 & 0.6899 & 0.6721 & 0.6950 \\
        4DGS2~\cite{yang2023gs4d} & \checkmark & 0.8247 & 0.7912 & 0.8271 & 0.8533 & 0.7546 & 0.6652 & 0.7738 & 0.7843 \\
        \midrule
        RoDynRF~\cite{liu2023robust} & - & 0.4255 & 0.4524 & 0.3810 & 0.5975 & 0.3621 & 0.3250 & 0.2914 & 0.4050 \\
        \ \ \ \ \ + TAM~\cite{yang2023track} & - & 0.3873 & 0.4504 & 0.2966 & 0.5500 & 0.2976 & 0.2879 & 0.3014 & 0.3673 \\
        \midrule
        RoDyGS (Ours) & - & 0.4347 & 0.5369 & 0.4627 & 0.5808 & 0.3921 & 0.4015 & 0.4502 & 0.4656 \\
        \bottomrule
    \end{tabular}
    }
    \vspace{-2mm}
    \caption{\textbf{SSIM results for iPhone dataset.}}
    \label{tab:ssim_iphone_all_results}
\end{table*}

\begin{table*}[!t]
    \centering
    \vspace{-2mm}
    \resizebox{1.0\linewidth}{!}{
    \begin{tabular}{lc|ccccccc|c}
    \toprule
     LPIPS($\downarrow$) & GT Cam & apple & block & paper-windmill & space-out & spin & teddy & wheel & average \\
        \midrule
        D-NeRF~\cite{pumarola2020d} & \checkmark & 0.4625 & 0.4957 & 0.4472 & 0.4452 & 0.5528 & 0.5433 & 0.5194 & 0.4952 \\
        RoDynRF~\cite{liu2023robust} & \checkmark & 0.5069 & 0.5702 & 0.4006 & 0.4167 & 0.4581 & 0.5675 & 0.5405 & 0.4943 \\
        4DGS1~\cite{wu20244d} & \checkmark & 0.4042 & 0.4031 & 0.3966 & 0.3469 & 0.4729 & 0.4276 & 0.4379 & 0.4142 \\
        Deform3D~\cite{yang2024deformable} & \checkmark & 0.3029 & 0.3069 & 0.3924 & 0.2420 & 0.3294 & 0.3033 & 0.2888 & 0.3094 \\
        4DGS2~\cite{yang2023gs4d} & \checkmark & 0.2448 & 0.2729 & 0.1569 & 0.2183 & 0.2860 & 0.3788 & 0.2166 & 0.2535 \\
        \midrule
        RoDynRF~\cite{liu2023robust} & - & 0.6022 & 0.5918 & 0.4244 & 0.5347 & 0.4950 & 0.6120 & 0.6114 & 0.5531 \\
        \ \ \ \ \ + TAM~\cite{yang2023track} & - & 0.6410 & 0.6282 & 0.5024 & 0.5299 & 0.5793 & 0.6744 & 0.5729 & 0.5897 \\
        \midrule
        RoDyGS (Ours) & - & 0.4850 & 0.4748 & 0.3569 & 0.4086 & 0.4278 & 0.4790 & 0.4203 & 0.4361 \\
        \bottomrule
    \end{tabular}
    }
    \vspace{-2mm}
    \caption{\textbf{LPIPS results for iPhone dataset.}}
    \label{tab:lpips_iphone_all_results}
\end{table*}

\begin{table*}[!t]
    \centering
    \vspace{-2mm}
    \resizebox{1.0\linewidth}{!}{
    \begin{tabular}{l|ccccccc|c}
    \toprule
     ATE($\downarrow$) & apple & block & paper-windmill & space-out & spin & teddy & wheel & average \\
        \midrule
        RoDynRF~\cite{liu2023robust} & 0.0312 & 0.0472 & 0.0124 & 0.0428 & 0.0050 & 0.0214 & 0.0348 & 0.0278 \\
        \ \ \ \ \ + TAM~\cite{yang2023track}& 0.0256 & 0.0435 & 0.0242 & 0.0411 & 0.0061 & 0.0359 & 0.0182 & 0.0278 \\
        \midrule
        RoDyGS (Ours) & 0.0044 & 0.0257 & 0.0065 & 0.0022 & 0.0060 & 0.0178 & 0.0246 & 0.0125 \\
        \bottomrule
    \end{tabular}
    }
    \vspace{-2mm}
    \caption{\textbf{
ATE results for iPhone dataset.}}
    \label{tab:ate_iphone_all}
\end{table*}

\begin{table*}[!t]
    \centering
    \vspace{-2mm}
    \resizebox{1.0\linewidth}{!}{
    \begin{tabular}{l|ccccccc|c}
    \toprule
     RPE-t($\downarrow$) & apple & block & paper-windmill & space-out & spin & teddy & wheel & average \\
        \midrule
        RoDynRF~\cite{liu2023robust} & 0.2784 & 0.1179 & 0.0911 & 0.2849 & 0.0610 & 0.1437 & 0.1585 & 0.1622 \\
        \ \ \ \ \ + TAM~\cite{yang2023track} & 0.2836 & 0.1097 & 0.1842 & 0.2803 & 0.0555 & 0.3656 & 0.1063 & 0.1979 \\
        \midrule
        RoDyGS (Ours) & 0.0358 & 0.4191 & 0.0315 & 0.0247 & 0.0820 & 0.1434 & 0.1293 & 0.1237 \\
        \bottomrule
    \end{tabular}
    }
    \vspace{-2mm}
    \caption{\textbf{
RPE-t results for iPhone dataset.}}
    \label{tab:rpe_t_iphone_all}
\end{table*}

\begin{table*}[!t]
    \centering
    \vspace{-2mm}
    \resizebox{1.0\linewidth}{!}{
    \begin{tabular}{l|ccccccc|c}
    \toprule
     RPE-R($\downarrow$) & apple & block & paper-windmill & space-out & spin & teddy & wheel & average \\
        \midrule
        RoDynRF~\cite{liu2023robust} & 0.4726 & 0.2501 & 0.1672 & 0.3325 & 0.1162 & 0.2827 & 0.3259 & 0.2782 \\
        \ \ \ \ \ + TAM~\cite{yang2023track}& 0.5154 & 0.2052 & 0.2222 & 0.3156 & 0.1157 & 0.3469 & 0.1776 & 0.2712 \\
        \midrule
        RoDyGS (Ours) & 0.1766 & 1.9176 & 0.1713 & 0.0247 & 0.2493 & 0.2662 & 0.3408 & 0.4495 \\
        \bottomrule
    \end{tabular}
    }
    \vspace{-2mm}
    \caption{\textbf{
RPE-R results for iPhone dataset.}}
    \label{tab:rpe_r_iphone_all}
\end{table*}

\begin{table*}[!h]
    \centering
    \begin{tabular}{l ccc| ccc| ccc}
        \toprule
        \multirow{2}{*}{iPhone~\cite{gao2022dycheck}} & \multicolumn{3}{c|}{RoDynRF + TAM~\cite{yang2023track}} & \multicolumn{3}{c|}{RoDynRF + MASt3R~\cite{leroy2024grounding}} & \multicolumn{3}{c}{RoDyGS (Ours)} \\
        \cmidrule(lr){2-4} \cmidrule(lr){5-7} \cmidrule(lr){8-10} 
        & PSNR(↑) & SSIM(↑) & LPIPS(↓) & PSNR(↑) & SSIM(↑) & LPIPS(↓) & PSNR(↑) & SSIM(↑) & LPIPS(↓) \\
        \hline
        apple & 12.99 & 0.3873 & 0.6410 & 16.51 & 0.4260 & 0.5523 & 16.79 & 0.4347 & 0.4850 \\
        block & 14.17 & 0.4504 & 0.6282 & 4.44 & 0.0030 & 0.7533 & 17.67 & 0.5369 & 0.4748 \\
        paper-windmill & 16.18 & 0.2966 & 0.5024 & 14.78 & 0.2184 & 0.5027 & 19.20 & 0.4627 & 0.3569 \\
        space-out & 16.84 & 0.5500 & 0.5299 & 16.06 & 0.5188 & 0.5541 & 19.16 & 0.5808 & 0.4086 \\
        spin & 13.80 & 0.2976 & 0.5793 & 14.55 & 0.2960 & 0.5751 & 18.47 & 0.3921 & 0.4278 \\
        wheel & 11.77 & 0.3014 & 0.5729 & 14.73 & 0.3526 & 0.6912 & 15.66 & 0.4502 & 0.4203 \\
        teddy & 11.35 & 0.2879 & 0.6744 & 15.79 & 0.3777 & 0.6036 & 14.69 & 0.4015 & 0.4790 \\
        \bottomrule  
    \end{tabular}
    
    \caption{\textbf{Scene-wise comparison between RoDynRF with off-the-shelf priors and RoDyGS.} We compare the dynamic novel view synthesis quality on the iPhone dataset between RoDyGS, RoDynRF initialized with MASt3R~\cite{leroy2024grounding} camera poses, and RoDynRF with TAM~\cite{yang2023track} masks.}
    \label{tab:ablation_scene_wise}
\end{table*}

\begin{table*}[!t]
    \centering
    \vspace{-2mm}
    \resizebox{1.0\linewidth}{!}{
    \begin{tabular}{lc|cccccccc|c}
    \toprule
     PSNR($\uparrow$) & GT Cam & scene0 & scene1 & scene2 & scene3 & scene4 & scene5 & scene6 & scene7 & average \\
        \midrule
        D-NeRF~\cite{pumarola2020d} & \checkmark & 21.95 & 19.90 & 22.73 & 20.56 & 23.62 & 20.61 & 19.11 & 16.68 & 20.65 \\
        RoDynRF~\cite{liu2023robust} & \checkmark & 23.19 & 21.44 & 22.82 & 20.56 & 22.39 & 20.74 & 18.00 & 17.30 & 20.80 \\
        4DGS1~\cite{wu20244d} &\checkmark & 24.68 & 23.38 & 24.95 & 21.76 & 23.92 & 22.65 & 20.19 & 18.13 & 22.46 \\
        Deform3D~\cite{yang2024deformable} &\checkmark & 24.31 & 23.57 & 25.51 & 22.12 & 24.15 & 22.44 & 20.68 & 18.50 & 22.66 \\
        4DGS2~\cite{yang2023gs4d} & \checkmark &25.29 & 24.03 & 24.86 & 21.99 & 23.99 & 23.00 & 22.07 & 18.89 & 23.02 \\
        \midrule
        RoDynRF ~\cite{liu2023robust} & - & 20.14 & 18.88 & 18.73 & 17.16 & 20.65 & 19.18 & 17.53 & 14.62 & 18.36 \\
        \ \ \ \ \ + TAM~\cite{yang2023track}  & -& 18.82 & 15.89 & 18.09 & 16.55 & 20.01 & 16.80 & 16.28 & 12.98 & 16.93 \\
        \midrule
        
        RoDyGS (Ours) & - &22.31 & 21.10 & 20.17 & 18.59 & 21.19 & 20.44 & 18.61 & 13.11 & 19.44
 \\
    \bottomrule
    \end{tabular}
    }
    \vspace{-2mm}
    \caption{\textbf{
PSNR results for Kubric-MRig dataset.}}
    \label{tab:psnr_kubric_mrig_all}
\end{table*}
\begin{table*}[!t]
    \centering
    \vspace{-2mm}
    \resizebox{1.0\linewidth}{!}{
    \begin{tabular}{lc|cccccccc|c}
    \toprule
     SSIM($\uparrow$) & GT Cam & scene0 & scene1 & scene2 & scene3 & scene4 & scene5 & scene6 & scene7 & average \\
        \midrule
        D-NeRF~\cite{pumarola2020d} & \checkmark & 0.7359 & 0.5533 & 0.7468 & 0.8069 & 0.7942 & 0.7299 & 0.7409 & 0.6202 & 0.7160 \\
        RoDynRF~\cite{liu2023robust} & \checkmark & 0.7773 & 0.6935 & 0.7854 & 0.8361 & 0.7299 & 0.8146 & 0.7707 & 0.6594 & 0.7584 \\
        4DGS1~\cite{wu20244d} & \checkmark & 0.8967 & 0.8802 & 0.9035 & 0.8563 & 0.8913 & 0.8639 & 0.7614 & 0.7924 & 0.8557 \\
        Deform3D~\cite{yang2024deformable} & \checkmark & 0.8178 & 0.8791 & 0.8998 & 0.8675 & 0.8857 & 0.8513 & 0.7803 & 0.8013 & 0.8478 \\
        4DGS2~\cite{yang2023gs4d} & \checkmark & 0.8695 & 0.8623 & 0.8846 & 0.8467 & 0.8806 & 0.8528 & 0.7856 & 0.7861 & 0.8460 \\
        \midrule
        RoDynRF~\cite{liu2023robust} & - & 0.6255 & 0.4772 & 0.6295 & 0.7153 & 0.6728 & 0.6958 & 0.6379 & 0.4432 & 0.6121 \\
        \ \ \ \ \ + TAM~\cite{yang2023track} & - & 0.6245 & 0.4457 & 0.6332 & 0.7351 & 0.6965 & 0.6522 & 0.6994 & 0.4393 & 0.6157 \\
        \midrule
        RoDyGS (Ours) & - & 0.7549 & 0.7043 & 0.7383 & 0.7667 & 0.7596 & 0.7810 & 0.6943 & 0.5360 & 0.7169 \\
        \bottomrule
    \end{tabular}
    }
    \vspace{-2mm}
    \caption{\textbf{SSIM results for Kubric-MRig dataset.}}
    \label{tab:ssim_kubric_mrig_all}
\end{table*}
\begin{table*}[!t]
    \centering
    \vspace{-2mm}
    \resizebox{1.0\linewidth}{!}{
    \begin{tabular}{lc|cccccccc|c}
    \toprule
     LPIPS ($\downarrow$) & GT Cam & scene0 & scene1 & scene2 & scene3 & scene4 & scene5 & scene6 & scene7 & average \\
        \midrule
        D-NeRF~\cite{pumarola2020d} & \checkmark & 0.4165 & 0.5112 & 0.3884 & 0.3357 & 0.3521 & 0.4543 & 0.3222 & 0.4364 & 0.4021 \\
        RoDynRF~\cite{liu2023robust} & \checkmark & 0.4443 & 0.4627 & 0.4244 & 0.4427 & 0.5494 & 0.5074 & 0.5936 & 0.4438 & 0.4836 \\
        4DGS1~\cite{wu20244d} & \checkmark & 0.1192 & 0.1279 & 0.1122 & 0.2054 & 0.1339 & 0.1935 & 0.3090 & 0.2073 & 0.1760 \\
        Deform3D~\cite{yang2024deformable} & \checkmark & 0.2013 & 0.1301 & 0.1300 & 0.1989 & 0.1445 & 0.2000 & 0.2974 & 0.1990 & 0.1877 \\
        4DGS2~\cite{yang2023gs4d} & \checkmark & 0.1563 & 0.1455 & 0.1407 & 0.2263 & 0.1457 & 0.2012 & 0.2860 & 0.2168 & 0.1898 \\
        \midrule
        RoDynRF~\cite{liu2023robust} & - & 0.5869 & 0.6555 & 0.5811 & 0.5357 & 0.5883 & 0.6062 & 0.6258 & 0.6081 & 0.5984 \\
        \ \ \ \ \ + TAM~\cite{yang2023track} & - & 0.6246 & 0.7292 & 0.6203 & 0.5343 & 0.5917 & 0.6120 & 0.6200 & 0.6472 & 0.6224 \\
        \midrule
        RoDyGS (Ours) & - & 0.2928 & 0.2818 & 0.3130 & 0.3413 & 0.2938 & 0.3149 & 0.4347 & 0.4481 & 0.3401 \\
        \bottomrule
    \end{tabular}
    }
    \vspace{-2mm}
    \caption{\textbf{LPIPS results for Kubric-MRig dataset.}}
    \label{tab:lpips_kubric_mrig_all}
\end{table*}

\begin{table*}[!t]
    \centering
    \vspace{-2mm}
    \resizebox{1.0\linewidth}{!}{
    \begin{tabular}{l|cccccccc|c}
    \toprule
     ATE($\downarrow$) & scene0 & scene1 & scene2 & scene3 & scene4 & scene5 & scene6 & scene7 & average \\
        \midrule
        RoDynRF~\cite{liu2023robust} & 0.0881 & 0.0468 & 0.0824 & 0.0724 & 0.0319 & 0.0603 & 0.0446 & 0.0793 & 0.0632 \\
        \ \ \ \ \ + TAM~\cite{yang2023track} & 0.0818 & 0.0693 & 0.0826 & 0.0764 & 0.0361 & 0.0801 & 0.0704 & 0.0593 & 0.0695 \\
        \midrule
        RoDyGS (Ours) & 0.0009 & 0.0025 & 0.0071 & 0.0061 & 0.0045 & 0.0088 & 0.0028 & 0.0085 & 0.0052 \\
        \bottomrule
    \end{tabular}
    }
    \vspace{-2mm}
    \caption{\textbf{
ATE results for Kubric-MRig dataset.}}
    \label{tab:ate_kubric_mrig_all}
\end{table*}

\begin{table*}[!t]
    \centering
    \vspace{-2mm}
    \resizebox{1.0\linewidth}{!}{
    \begin{tabular}{l|cccccccc|c}
    \toprule
     RPE-t($\downarrow$) & scene0 & scene1 & scene2 & scene3 & scene4 & scene5 & scene6 & scene7 & average \\
        \midrule
        RoDynRF~\cite{liu2023robust} & 0.5425 & 0.3475 & 0.5030 & 0.4713 & 0.2174 & 0.3328 & 0.3190 & 0.5368 & 0.4088 \\
        \ \ \ \ \ + TAM~\cite{yang2023track}& 0.7732 & 0.4195 & 0.4228 & 0.6579 & 0.2550 & 0.4593 & 0.4750 & 0.2982 & 0.4701 \\
        \midrule
        RoDyGS (Ours) & 0.0083 & 0.0207 & 0.0563 & 0.0512 & 0.0320 & 0.0479 & 0.0223 & 0.0601 & 0.0374 \\
        \bottomrule
    \end{tabular}
    }
    \vspace{-2mm}
    \caption{\textbf{
RPE-t results for Kubric-MRig dataset.}}
    \label{tab:rpe_t_kubric_mrig_all}
\end{table*}

\begin{table*}[!t]
    \centering
    \vspace{-2mm}
    \resizebox{1.0\linewidth}{!}{
    \begin{tabular}{l|cccccccc|c}
    \toprule
     RPE-R($\downarrow$) & scene0 & scene1 & scene2 & scene3 & scene4 & scene5 & scene6 & scene7 & average \\
        \midrule
        RoDynRF~\cite{liu2023robust} & 1.9904 & 1.6781 & 2.1122 & 1.7471 & 1.5009 & 1.9715 & 1.7836 & 1.8201 & 1.8255 \\
        \ \ \ \ \ + TAM~\cite{yang2023track} & 2.2493 & 1.7614 & 2.4783 & 1.8486 & 1.5099 & 1.8526 & 1.8043 & 1.4920 & 1.8746 \\
        \midrule
        RoDyGS (Ours) & 0.0187 & 0.0739 & 0.1004 & 0.1170 & 0.0902 & 0.1494 & 0.1036 & 0.0933 & 0.0933 \\
        \bottomrule
    \end{tabular}
    }
    \vspace{-2mm}
    \caption{\textbf{
RPE-R results for Kubric-MRig dataset.}}
    \label{tab:rpe_r_kubric_mrig_all}
\end{table*}

\begin{table*}[!t]
    \centering
    \vspace{-2mm}
    \resizebox{1.0\linewidth}{!}{
    \begin{tabular}{lc|ccccccc|c}
    \toprule
     PSNR($\uparrow$) & GT Cam & apple & block & paper-windmill & space-out & spin & teddy & wheel & average \\
        \midrule
        D-NeRF~\cite{pumarola2020d} & \checkmark & 24.23 & 21.80 & 21.85 & 25.18 & 22.15 & 19.46 & 19.65 & 22.04 \\
        RoDynRF~\cite{liu2023robust} & \checkmark & 17.38 & 15.99 & 20.71 & 20.62 & 16.66 & 13.28 & 12.78 & 16.78 \\
        4DGS1~\cite{wu20244d} & \checkmark & 23.24 & 22.05 & 21.03 & 24.81 & 22.99 & 18.89 & 17.72 & 21.25 \\
        Deform3D~\cite{yang2024deformable} & \checkmark & 24.82 & 23.26 & 20.62 & 26.95 & 23.51 & 20.93 & 19.56 & 22.81 \\
        4DGS2~\cite{yang2023gs4d} & \checkmark & 27.31 & 25.09 & 26.81 & 29.54 & 26.30 & 21.11 & 22.45 & 25.52 \\
        \midrule
        RoDynRF~\cite{liu2023robust} & - & 14.50 & 14.73 & 17.94 & 18.06 & 15.75 & 11.56 & 11.85 & 14.91 \\
        \ \ \ \ \ + TAM~\cite{yang2023track} & - & 12.99 & 14.17 & 16.18 & 16.84 & 13.80 & 11.35 & 11.77 & 13.87 \\
        \midrule
        RoDyGS (Ours) & - & 16.79 & 17.67 & 19.20 & 19.16 & 18.47 & 14.69 & 15.66 & 17.38 \\
    \bottomrule
    \end{tabular}
    }
    \vspace{-2mm}
    \caption{\textbf{PSNR results for iPhone dataset.}}
    \label{tab:psnr_iphone_all_results}
\end{table*}

\begin{table*}[!t]
    \centering
    \vspace{-2mm}
    \resizebox{1.0\linewidth}{!}{
    \begin{tabular}{lc|ccccccc|c}
    \toprule
     SSIM($\uparrow$) & GT Cam & apple & block & paper-windmill & space-out & spin & teddy & wheel & average \\
        \midrule
        D-NeRF~\cite{pumarola2020d} & \checkmark & 0.6748 & 0.6238 & 0.5129 & 0.7018 & 0.5002 & 0.4940 & 0.5665 & 0.5820 \\
        RoDynRF~\cite{liu2023robust} & \checkmark & 0.5310 & 0.4858 & 0.5760 & 0.6844 & 0.5139 & 0.3806 & 0.3959 & 0.5096 \\
        4DGS1~\cite{wu20244d} & \checkmark & 0.6512 & 0.6654 & 0.5080 & 0.7385 & 0.5283 & 0.5562 & 0.5224 & 0.5865 \\
        Deform3D~\cite{yang2024deformable} & \checkmark & 0.7394 & 0.7377 & 0.5557 & 0.8177 & 0.6528 & 0.6899 & 0.6721 & 0.6950 \\
        4DGS2~\cite{yang2023gs4d} & \checkmark & 0.8247 & 0.7912 & 0.8271 & 0.8533 & 0.7546 & 0.6652 & 0.7738 & 0.7843 \\
        \midrule
        RoDynRF~\cite{liu2023robust} & - & 0.4255 & 0.4524 & 0.3810 & 0.5975 & 0.3621 & 0.3250 & 0.2914 & 0.4050 \\
        \ \ \ \ \ + TAM~\cite{yang2023track} & - & 0.3873 & 0.4504 & 0.2966 & 0.5500 & 0.2976 & 0.2879 & 0.3014 & 0.3673 \\
        \midrule
        RoDyGS (Ours) & - & 0.4347 & 0.5369 & 0.4627 & 0.5808 & 0.3921 & 0.4015 & 0.4502 & 0.4656 \\
        \bottomrule
    \end{tabular}
    }
    \vspace{-2mm}
    \caption{\textbf{SSIM results for iPhone dataset.}}
    \label{tab:ssim_iphone_all_results}
\end{table*}

\begin{table*}[!t]
    \centering
    \vspace{-2mm}
    \resizebox{1.0\linewidth}{!}{
    \begin{tabular}{lc|ccccccc|c}
    \toprule
     LPIPS($\downarrow$) & GT Cam & apple & block & paper-windmill & space-out & spin & teddy & wheel & average \\
        \midrule
        D-NeRF~\cite{pumarola2020d} & \checkmark & 0.4625 & 0.4957 & 0.4472 & 0.4452 & 0.5528 & 0.5433 & 0.5194 & 0.4952 \\
        RoDynRF~\cite{liu2023robust} & \checkmark & 0.5069 & 0.5702 & 0.4006 & 0.4167 & 0.4581 & 0.5675 & 0.5405 & 0.4943 \\
        4DGS1~\cite{wu20244d} & \checkmark & 0.4042 & 0.4031 & 0.3966 & 0.3469 & 0.4729 & 0.4276 & 0.4379 & 0.4142 \\
        Deform3D~\cite{yang2024deformable} & \checkmark & 0.3029 & 0.3069 & 0.3924 & 0.2420 & 0.3294 & 0.3033 & 0.2888 & 0.3094 \\
        4DGS2~\cite{yang2023gs4d} & \checkmark & 0.2448 & 0.2729 & 0.1569 & 0.2183 & 0.2860 & 0.3788 & 0.2166 & 0.2535 \\
        \midrule
        RoDynRF~\cite{liu2023robust} & - & 0.6022 & 0.5918 & 0.4244 & 0.5347 & 0.4950 & 0.6120 & 0.6114 & 0.5531 \\
        \ \ \ \ \ + TAM~\cite{yang2023track} & - & 0.6410 & 0.6282 & 0.5024 & 0.5299 & 0.5793 & 0.6744 & 0.5729 & 0.5897 \\
        \midrule
        RoDyGS (Ours) & - & 0.4850 & 0.4748 & 0.3569 & 0.4086 & 0.4278 & 0.4790 & 0.4203 & 0.4361 \\
        \bottomrule
    \end{tabular}
    }
    \vspace{-2mm}
    \caption{\textbf{LPIPS results for iPhone dataset.}}
    \label{tab:lpips_iphone_all_results}
\end{table*}

\begin{table*}[!t]
    \centering
    \vspace{-2mm}
    \resizebox{1.0\linewidth}{!}{
    \begin{tabular}{l|ccccccc|c}
    \toprule
     ATE($\downarrow$) & apple & block & paper-windmill & space-out & spin & teddy & wheel & average \\
        \midrule
        RoDynRF~\cite{liu2023robust} & 0.0312 & 0.0472 & 0.0124 & 0.0428 & 0.0050 & 0.0214 & 0.0348 & 0.0278 \\
        \ \ \ \ \ + TAM~\cite{yang2023track}& 0.0256 & 0.0435 & 0.0242 & 0.0411 & 0.0061 & 0.0359 & 0.0182 & 0.0278 \\
        \midrule
        RoDyGS (Ours) & 0.0044 & 0.0257 & 0.0065 & 0.0022 & 0.0060 & 0.0178 & 0.0246 & 0.0125 \\
        \bottomrule
    \end{tabular}
    }
    \vspace{-2mm}
    \caption{\textbf{
ATE results for iPhone dataset.}}
    \label{tab:ate_iphone_all}
\end{table*}

\begin{table*}[!t]
    \centering
    \vspace{-2mm}
    \resizebox{1.0\linewidth}{!}{
    \begin{tabular}{l|ccccccc|c}
    \toprule
     RPE-t($\downarrow$) & apple & block & paper-windmill & space-out & spin & teddy & wheel & average \\
        \midrule
        RoDynRF~\cite{liu2023robust} & 0.2784 & 0.1179 & 0.0911 & 0.2849 & 0.0610 & 0.1437 & 0.1585 & 0.1622 \\
        \ \ \ \ \ + TAM~\cite{yang2023track} & 0.2836 & 0.1097 & 0.1842 & 0.2803 & 0.0555 & 0.3656 & 0.1063 & 0.1979 \\
        \midrule
        RoDyGS (Ours) & 0.0358 & 0.4191 & 0.0315 & 0.0247 & 0.0820 & 0.1434 & 0.1293 & 0.1237 \\
        \bottomrule
    \end{tabular}
    }
    \vspace{-2mm}
    \caption{\textbf{
RPE-t results for iPhone dataset.}}
    \label{tab:rpe_t_iphone_all}
\end{table*}

\begin{table*}[!t]
    \centering
    \vspace{-2mm}
    \resizebox{1.0\linewidth}{!}{
    \begin{tabular}{l|ccccccc|c}
    \toprule
     RPE-R($\downarrow$) & apple & block & paper-windmill & space-out & spin & teddy & wheel & average \\
        \midrule
        RoDynRF~\cite{liu2023robust} & 0.4726 & 0.2501 & 0.1672 & 0.3325 & 0.1162 & 0.2827 & 0.3259 & 0.2782 \\
        \ \ \ \ \ + TAM~\cite{yang2023track}& 0.5154 & 0.2052 & 0.2222 & 0.3156 & 0.1157 & 0.3469 & 0.1776 & 0.2712 \\
        \midrule
        RoDyGS (Ours) & 0.1766 & 1.9176 & 0.1713 & 0.0247 & 0.2493 & 0.2662 & 0.3408 & 0.4495 \\
        \bottomrule
    \end{tabular}
    }
    \vspace{-2mm}
    \caption{\textbf{
RPE-R results for iPhone dataset.}}
    \label{tab:rpe_r_iphone_all}
\end{table*}

\end{document}